\documentclass{article} % For LaTeX2e
\usepackage{iclr2027_conference,times}
\iclrfinalcopy
\usepackage{amsmath,amsfonts,bm}

\def\eqref#1{equation~\ref{#1}}
\def\1{\bm{1}}

\DeclareMathAlphabet{\mathsfit}{\encodingdefault}{\sfdefault}{m}{sl}
\SetMathAlphabet{\mathsfit}{bold}{\encodingdefault}{\sfdefault}{bx}{n}

\usepackage{hyperref}
\hypersetup{colorlinks=true,linkcolor=blue,citecolor=blue,urlcolor=blue}
\usepackage{etoc}
\usepackage{url}
\usepackage{graphicx}
\usepackage{booktabs}
\usepackage{wrapfig}
\usepackage[labelformat=simple]{subcaption} % phantom sub-labels so \ref gives e.g. 3a

\usepackage{multirow}
\usepackage{pgfplots}
\usepackage{adjustbox}
\usepgfplotslibrary{groupplots,colormaps}
\pgfplotsset{compat=1.18}
\usetikzlibrary{positioning,arrows.meta}
\usetikzlibrary{calc}
\DeclareRobustCommand{\upglyph}{\textcolor{green!55!black}{\tikz[baseline=0pt]\fill (0,0.1ex)--(0.9ex,0.1ex)--(0.45ex,0.95ex)--cycle;}}
\DeclareRobustCommand{\dnglyph}{\textcolor{red!80!black}{\tikz[baseline=0pt]\fill (0,0.95ex)--(0.9ex,0.95ex)--(0.45ex,0.1ex)--cycle;}}
\usepackage{xcolor}
\usepackage{algorithm}
\usepackage{algpseudocode}
\usepackage{tabularx}
\usepackage{array}
\usepackage{makecell}

\newcommand{\cW}{\mathcal{W}}
\newcommand{\bw}{\bm{w}}
\newcommand{\ba}{\bm{a}}
\newcommand{\pidp}{\pi_{\mathrm{dp}}}
\newcommand{\pidpw}{\pi_{\mathrm{dp}}^{\scriptscriptstyle \cW}}
\newcommand{\piw}{\pi^{\scriptscriptstyle \cW}}
\newcommand{\Mwm}{\mathcal{M}_{\mathrm{wm}}}
\newcommand{\Menv}{\mathcal{M}_{\mathrm{env}}}

\title{Dream to Remember: Continual Steering of Generalist Robot Policies with World Models}

\author{
\makebox[\textwidth][c]{%
\textbf{Bahey Tharwat}$^{*1}$ \quad
\textbf{Iman Nematollahi}$^{*1}$ \quad
\textbf{Marvin Alles}$^{2}$ \quad
\textbf{Xingyuan Zhang}$^{2}$
}\\[4pt]
\makebox[\textwidth][c]{%
\textbf{Akshay L Chandra}$^{1}$ \quad
\textbf{Abhinav Valada}$^{1}$
}\\[8pt]
\makebox[\textwidth][c]{%
\normalfont
$^{1}$University of Freiburg \qquad
$^{2}$Technical University of Munich
}
}

\begin{document}
\etocdepthtag.toc{mtappendix}

\maketitle
\lhead{Preprint.}
\let\thefootnote\relax
\footnotetext{$^{*}$Equal contribution. Project page: \href{https://d2r.cs.uni-freiburg.de}{d2r.cs.uni-freiburg.de}}
\setcounter{footnote}{0}

\begin{abstract}

Generalist robot policies are trained with imitation learning and are bounded by the quality and coverage of their demonstrations. Reinforcement learning can improve and extend them, either by fine-tuning the policy weights or by training a small module to steer the input noise of a frozen generative policy. 
When tasks arrive sequentially, both approaches suffer from catastrophic forgetting. Fine-tuning on a task degrades prior tasks, and steering only shifts forgetting to the shared steering module.
We propose Dream to Remember (D2R), which continually steers a frozen generalist policy toward new tasks, while retaining its earlier skills. As the steering module learns a new task, a frozen pretrained world model rehearses prior tasks through imagined rollouts, anchoring the module to its earlier behavior without a replay buffer or new data. When the world model is faithful enough, D2R learns the new task entirely in imagination. Otherwise, it learns the task in the environment while using the world model only to rehearse prior tasks. On CALVIN, D2R improves eight tasks in sequence entirely in imagination, raising average success from 63.7\% to 74.8\%, while LoRA fine-tuning on the same sequence drops it to 11.7\%. With a frozen off-the-shelf generalist policy and world model, $\pi_0$ and A2World, D2R learns unseen LIBERO-90 tasks in the environment while keeping $\pi_0$'s performance on the LIBERO suites it was fine-tuned on, even though the world model is not accurate enough for policy optimization. Finally, on a real Franka robot, D2R improves three tasks in sequence entirely in imagination while retaining the earlier improvements.

\end{abstract}

%%%%%%%%%%%%%%%%%%%%%%%%%%%%%%%%%%%%%%%%%%%%%%%%%%%%%%%%%%%%%%%%%%%%%%%%%%%%%%%%
\section{INTRODUCTION}

A robot operating in a home or warehouse should continually improve existing skills, acquire new ones, and retain what it has already learned. However, even today's generalist robot policies trained on large demonstration corpora \citep{chi2023diffusionpolicy,lipman2023flowmatching,black2024pi_0,bjorck2025grootn1} do not natively support continual adaptation that reliably preserves prior skills. Imitation learning limits these policies to demonstrated behavior, including its flaws and gaps, but those same demonstrations provide a strong prior worth preserving~\citep{von2024art,heppert2026scaling}. Reinforcement learning can improve and extend that prior through the robot's own experience, whether gathered on the real robot \citep{intelligence2025pi,dong2026expo}, in simulation \citep{chen2025pirl}, or through imagined rollouts of a learned world model \citep{chandra2025diwa,sharma2026world}, but doing so without catastrophic forgetting remains a central challenge.\looseness=-1

Where adaptation occurs determines where forgetting can happen. Fine-tuning places it in the policy itself, so learning one task can alter behavior on others, whether the update modifies all weights or only an adapter \citep{hu2022lora,zhu2026continualvla}. Test-time action-selection methods sidestep this by adapting nothing persistent. They keep the policy frozen, sample multiple action proposals, score them against an external evaluator, such as a world model or a critic, then execute the best one \citep{nakamoto2025vgps,cui2026dreamsteer,qi2025gpc}, but pay this significant search cost at every decision. Input steering keeps the policy frozen while amortizing the search into a learned module. Since a diffusion policy maps noise to actions, the module can learn which noise to sample to steer the policy toward better behavior \citep{wagenmaker2025dsrl}. Steering, however, only moves forgetting to the shared steering network. Training it on a new task changes the noise sampled for previous ones, so forgetting reappears one level down.\looseness=-1

\begin{figure}[t]
    \centering
    \includegraphics[width=\textwidth]{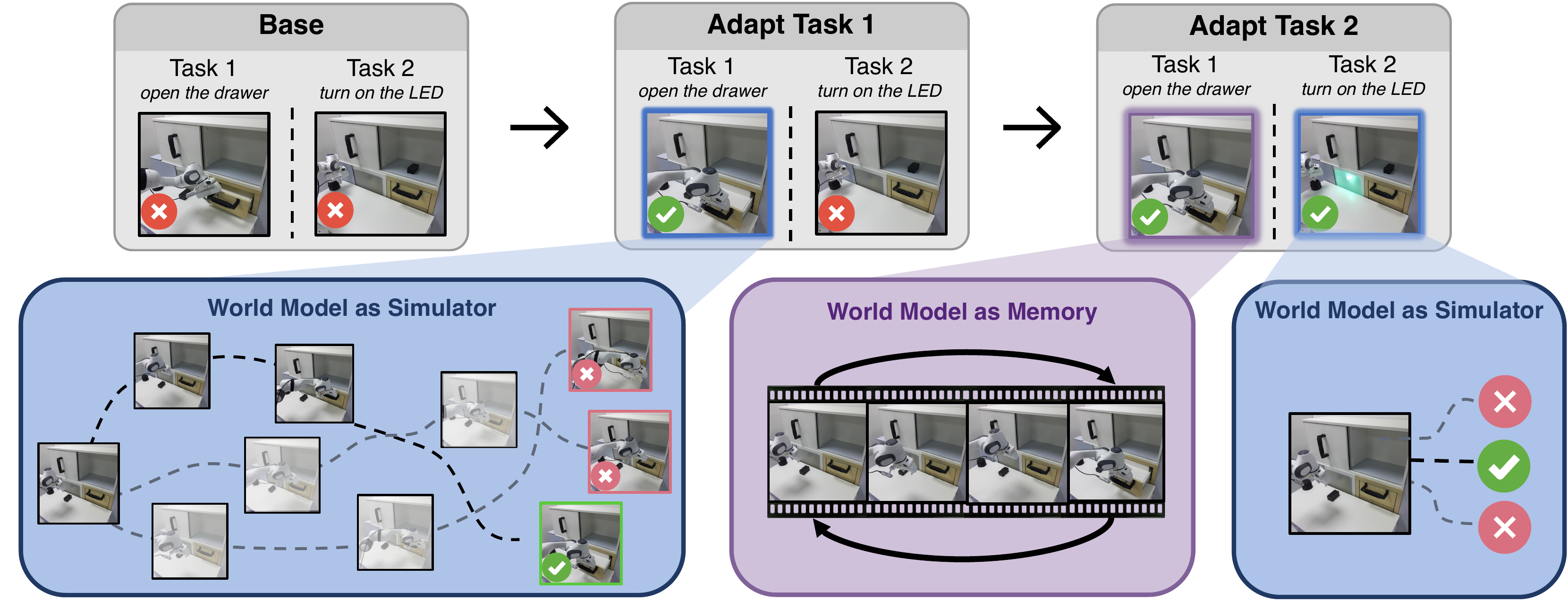}
    \caption{\textbf{Dream to Remember (D2R)}: We adapt a frozen generative policy sequentially by steering its input noise with a small module trained inside a learned world model. While the module learns the current task, the same world model regenerates imagined rollouts of prior tasks, and we regularize the module toward its previous behavior on those states. Skills thus accumulate without revisiting prior tasks on the robot or storing their trajectories. Rehearsal requires less world model fidelity than optimization, allowing even a model too coarse for learning a new task to preserve prior skills.}
    \vspace{-0.5cm}
    \label{fig:teaser}
\end{figure}

A common way to mitigate forgetting is rehearsal, where past tasks are revisited while learning a new one \citep{rolnick2019experience}. Rehearsal can come from fresh robot interaction, at the cost of rerunning every past task in each round of adaptation, or from experience replay, which stores past trajectories \citep{buzzega2020dark}, requiring a buffer that grows over time \citep{du2026memory}.

We propose \textbf{Dream to Remember (D2R)}, which adapts a frozen generative policy through input steering and prevents the steering module from forgetting by rehearsing prior tasks in a learned world model. For the current task, whether refining or acquiring a skill, the steering module learns from real experience or, when the world model is sufficiently faithful, entirely from imagined rollouts. For prior tasks, the same world model provides rehearsal. Starting from an initial observation of each task, it regenerates rollouts under the steering behavior to be preserved, and the current module is regularized to match that behavior on the imagined states. Since the world model is frozen, the rehearsal distribution does not drift as new tasks arrive, and regenerating rollouts avoids costly repeated robot interaction and storing past trajectories. Importantly, we show that rehearsal requires less from the world model than learning a new task. Learning needs reliable predictions wherever the module explores for reward, whereas rehearsal only requires familiar states to be reproduced closely enough to anchor the module. D2R can therefore use a frozen, off-the-shelf world model never trained on the target tasks, too inaccurate for learning, yet sufficient for rehearsal.

We evaluate D2R in three settings with different levels of world model fidelity. On CALVIN \citep{mees2022calvin}, where a world model trained on the robot's own play data is sufficiently faithful for policy optimization, D2R improves eight tasks sequentially and entirely in imagination, raising average success from 63.7\% to 74.8\%. Over the same sequence, LoRA fine-tuning reduces this average to 11.7\%, while steering without rehearsal leaves it at the base level. On LIBERO-90, we instead use a frozen generalist policy, $\pi_0$ \citep{black2024pi_0}, and a frozen pretrained world model, A2World \citep{huang2026learning}, neither trained on the target tasks. Here, policies cloned from world model rollouts perform far worse than those cloned from the same real demonstrations, showing that the model is not accurate enough for policy optimization. Yet it remains effective for rehearsal, enabling D2R to learn unseen tasks through environment interaction while preserving $\pi_0$'s performance on the LIBERO suites on which it was previously fine-tuned, ending 1.6 points above its original success, whereas steering without rehearsal loses 30.8 points. Finally, on a real Franka robot, D2R sequentially improves three tasks using only imagined rollouts while retaining the gains on them.\looseness=-1

In summary, our contributions are threefold. (i)~\textbf{Steering in imagination.} We train the steering module of a frozen policy inside a learned world model, matching policy fine-tuning on the adapted task while retaining other skills far better. (ii)~\textbf{World model as memory.} We use a frozen world model to rehearse prior tasks by regenerating their rollouts on demand, enabling continual steering without repeated robot interaction or stored trajectories. (iii)~\textbf{World-model rehearsal needs less fidelity than optimization.} We show that an off-the-shelf world model can be inaccurate for policy optimization yet sufficient for rehearsal, enabling continual adaptation with frozen $\pi_0$ and A2World. We make our code publicly available to support reproducibility.

\section{Related Work}

\textbf{Improving pretrained generative policies.}~Prior work primarily adapts a pretrained generative policy to a single task. In weight space, reinforcement learning can update all weights through the denoising process \citep{ren2025diffusion,chen2025pirl} or be restricted to low-rank adapters \citep{hu2022lora}. DiWA \citep{chandra2025diwa} performs such fine-tuning inside a frozen latent world model \citep{hafner2020dreamer,demoss2023ditto,nematollahi2025lumos}, while generative video models have also been used as simulators for VLA fine-tuning \citep{xiao2025worldenv,li2025vlarft,sharma2026world}. In output space, methods either rank and select from sampled actions \citep{nakamoto2025vgps,qi2025gpc,cui2026dreamsteer,kwok2025robomonkey} or learn bounded action residuals \citep{dong2026expo}. In input space, DSRL \citep{wagenmaker2025dsrl} learns the sampling noise with reinforcement learning on the robot, SCORE \citep{yu2026score} learns it in simulation and transfers to the real robot, and FlowDAgger \citep{murray2026flowdagger} finds that noise-space adaptation preserves held-out skills better than full fine-tuning, LoRA, or action residuals. These methods focus on individual-task adaptation. D2R instead carries noise-space steering across tasks and rehearses prior tasks in the world model, even when it is too coarse for policy optimization.

\textbf{Continual learning without forgetting.}~Sequential learning requires balancing stability and plasticity \citep{kirkpatrick2017ewc,dohare2024loss,khetarpal2022continualrl, raghavan2024flexible}. Regularization limits changes to parameters or outputs important for earlier tasks \citep{kirkpatrick2017ewc,zenke2017si,li2017lwf}, parameter-isolation methods allocate separate capacity across tasks \citep{rusu2016progressive,mallya2018packnet}, and rehearsal revisits stored or generated past data \citep{lopezpaz2017gem,rolnick2019experience,buzzega2020dark,shin2017dgr,yue2024tdgr}. For robot policies, TAIL \citep{liu2024tail} adds task-specific adapters, CLARE and CORAL \citep{romer2026clare,liu2026coral} expand lightweight modules with routing, and LOTUS \citep{wan2024lotus} grows a library of skill policies, all reducing forgetting at the cost of model growth with each task. Forgetting also depends on the adaptation signal. On-policy RL through low-rank adapters can retain prior VLA skills well \citep{hu2026continualvlarl}, consistent with broader findings that RL \citep{shenfeld2025razor} and LoRA \citep{biderman2024lora} can reduce forgetting, whereas sequential supervised fine-tuning shows far more forgetting and benefits from replay \citep{he2026cpc,zhu2026continualvla}. This retention relies on small updates within a large pretrained network and requires access to its weights, gradients, and on-policy interaction. D2R instead updates only a small steering module with forward access to the policy and uses a separate frozen world model for memory, keeping a single module without task identities and storing only initial observations of retained tasks.\looseness=-1

%%%%%%%%%%%%%%%%%%%%%%%%%%%%%%%%%%%%%%%%%%%%%%%%%%%%%%%%%%%%%%%%%%%%%%%%%%%%%%%%
\section{Method}
\label{sec:method}
\begin{figure}[t]
    \centering
    \includegraphics[width=\textwidth]{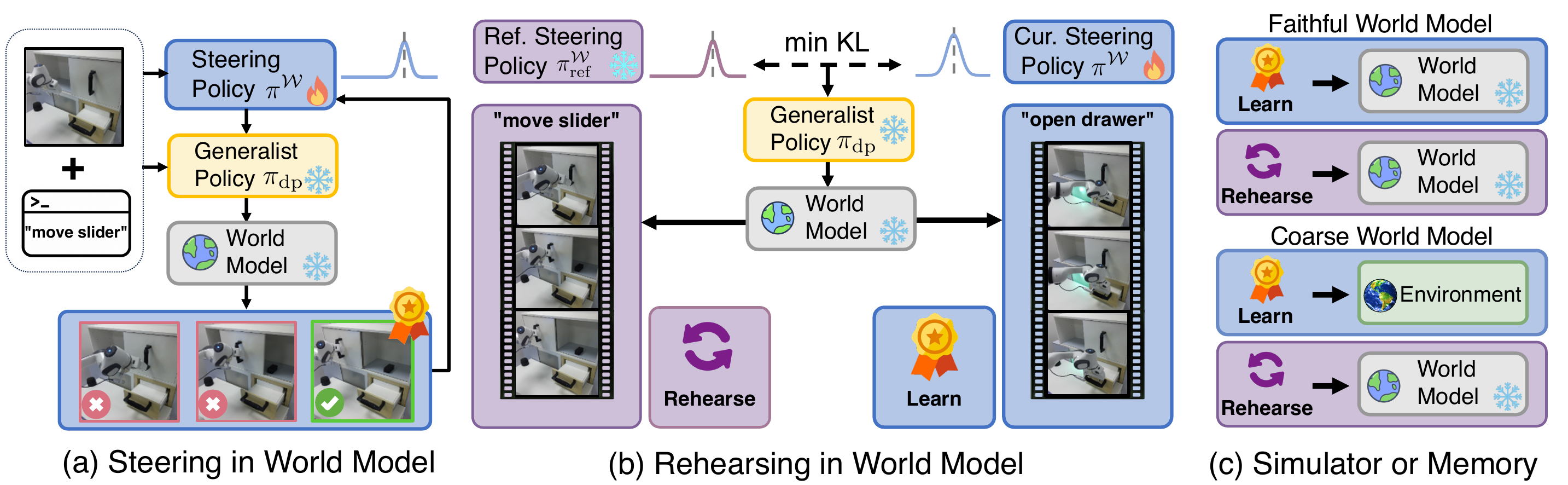}
    \caption{\textbf{Overview of D2R.} (a)~\textbf{Steering in imagination:} A small steering module learns to shape the noise fed to the frozen policy through RL in a frozen world model. (b)~\textbf{Rehearsal in the world model:} The world model imagines rollouts of prior tasks, where a KL penalty anchors the module to its earlier behavior. (c)~\textbf{Simulator or memory:} A faithful world model supports both learning and rehearsal, while a coarser one can still support rehearsal as learning moves to the environment.}
    \vspace{-0.2cm}
    \label{fig:method}
\end{figure}

In this section, we present D2R, illustrated in Figure~\ref{fig:method}. We first define the continual setting, where tasks arrive sequentially and previously learned tasks must be retained (\S\ref{sec:problem_formulation}). We then describe how a small steering module adapts the frozen policy to the current task using imagined rollouts (\S\ref{sec:steering}), how world model rehearsal keeps the module from forgetting the other tasks (\S\ref{sec:rehearsal}), and how a world model that is not accurate enough for policy optimization can still be used for rehearsal (\S\ref{sec:two_roles}). %Algorithm~\ref{alg:d2r} in the appendix summarizes the full procedure.

\subsection{Problem Formulation}
\label{sec:problem_formulation}

\textbf{Continual adaptation.}~We consider a robot with a pretrained generalist policy adapting to a sequence of language-conditioned tasks $\ell_1,\dots,\ell_N$ that arrive one at a time. At stage $k$, the robot can gather experience only on $\ell_k$, which may refine an existing skill or learn a new one. The goal is to improve success on $\ell_k$ while preserving the robot's existing skills, which form the \emph{retention set} $\mathcal{R}_k$. Our goal is to preserve these skills without revisiting them on the robot or storing their trajectories, while assuming access only to a small set of initial observations per task.

\textbf{Frozen generative policy.}~Robotic control is naturally modelled as a visual POMDP, where the agent receives high-dimensional visual observations $o \in \mathcal{O}$ and task-dependent rewards $R_k$. To retain standard MDP notation, we define the environment as $\Menv=(\mathcal{S},\mathcal{A},P,R_k,\gamma)$, where $s \in \mathcal{S}$ denotes the Markovian state representation constructed from the observation and context history. The robot is controlled by a frozen generative policy $\pidp$, a diffusion or flow-matching model pretrained by imitation on multi-task demonstrations, that maps Gaussian noise $\bw\sim\mathcal{N}(0,I)$, with $\bw\in\cW=\mathbb{R}^{H\cdot d}$, to an action chunk $\ba\in\mathcal{A}^H$ conditioned on the state $s$ and instruction $\ell$. Under deterministic sampling, such as DDIM for diffusion \citep{song2020denoising} or ODE integration for flow matching \citep{lipman2023flowmatching}, the action is a deterministic function of the initial noise and conditioning, denoted as $\ba=\pidpw(s,\ell,\bw)$. We require only forward access to $\pidpw$, not its weights or gradients, so any off-the-shelf policy can be used.

\textbf{World model.}~A learned world model simulates environment dynamics over state representations $s \in \mathcal{S}$. Concretely, an encoder $E_\phi$ maps observation histories to states $s_t = E_\phi(o_{\le t})$, and a transition model $P_\phi(s' \mid s,\ba)$ predicts the next latent state from an action chunk \citep{hafner2020dreamer,huang2026learning}. Replacing the environment dynamics with $P_\phi$ defines the world model MDP $\Mwm=(\mathcal{S},\mathcal{A},P_\phi,R_k,\gamma)$, which shares the action space and reward with $\Menv$ but can be rolled out without environment interaction. In imagination, a learned success classifier evaluates $R_k$ on latent states. We assume the world model is given and remains frozen throughout, whether trained on the robot's own data or pretrained as a generalist model without exposure to the target tasks.\looseness=-1

\textbf{Objective.}~Given the frozen policy and world model, we learn a \emph{steering policy} $\piw(\bw\mid s,\ell)$ that maximizes return on the current task while preserving behavior on the retention set. Formally,
\begin{equation}
\label{eq:objective}
\max_{\piw}\ J_k(\piw)
\qquad \text{s.t.} \qquad
\text{KL}(\piw(\cdot\mid s,\ell_j) \| \piw_{\mathrm{ref}(j)}(\cdot\mid s,\ell_j)) \leq \epsilon
\quad \forall\, \ell_j\in\mathcal{R}_k,
\end{equation}
where $J_k$ is the expected return on $\ell_k$, $\smash{\piw_{\mathrm{ref}(j)}}$ is the steering behavior to preserve for task $\ell_j$, and $\epsilon$ bounds its allowed deviation, while adaptation updates only $\piw$. Section~\ref{sec:steering} describes how D2R maximizes return via steering in imagination, while Section~\ref{sec:rehearsal} enforces the retention constraints.\looseness=-1

\subsection{Steering a Frozen Policy in Imagination}
\label{sec:steering}

We first address the return term in Eq.~\ref{eq:objective}, improving the frozen policy on the current task without updating its weights and, when possible, without environment interaction. Since $\pidpw$ is fixed, choosing a noise vector $\bw$ induces an action, allowing the noise space to serve as the action space of a transformed MDP. We define $P^{\cW}(s' \mid s,\ell,\bw)=P\big(s' \mid s,\pidpw(s,\ell,\bw)\big)$ and $R^{\cW}_k(s,\ell,\bw)=R_k\big(s,\pidpw(s,\ell,\bw)\big)$. The resulting \emph{latent-noise MDP} $\Menv^{\cW}=(\mathcal{S},\cW,P^{\cW},R^{\cW}_k,\gamma)$ treats the frozen policy as part of the environment, allowing RL to optimize directly over $\cW$. DSRL \citep{wagenmaker2025dsrl} learns $\piw$ in this MDP through interactions with the real environment. We instead replace the environment dynamics with the world model, yielding the \emph{imagined latent-noise MDP}\looseness=-1 
\begin{equation}
\label{eq:imagined_mdp}
\Mwm^{\cW} = \big(\mathcal{S},\cW,P^{\cW}_\phi,R^{\cW}_k,\gamma\big),
\qquad
P^{\cW}_\phi(s' \mid s,\ell,\bw) = P_\phi\big(s' \mid s,\pidpw(s,\ell,\bw)\big).
\end{equation}
When the world model is sufficiently faithful, we train $\piw$ in $\Mwm^{\cW}$ using only forward passes through the frozen policy and world model, with no environment interaction. Otherwise, we train it in $\Menv^{\cW}$, as discussed in \S\ref{sec:two_roles}. In either case, the current task is learned by minimizing the loss $\mathcal{L}_{\mathrm{RL}}$ of a standard RL algorithm over $\cW$ (see Appendix, Eqs.~\ref{eq:app_ppo}, \ref{eq:app_sac}). A small network implements the steering module $\piw$ which is orders of magnitude smaller than $\pidp$ and maps the state $s$ and instruction $\ell$ to a diagonal Gaussian over the noise,\looseness=-1
\begin{equation}
\label{eq:steer}
\piw(\bw \mid s,\ell) = \mathcal{N}\big(\bw;\ \mu(s,\ell),\ \mathrm{diag}\,\sigma^2(s,\ell)\big),
\qquad \sigma\in[\sigma_{\min},\sigma_{\max}].
\end{equation}
Setting $\mu=0$ and $\sigma=1$ recovers the original noise distribution, allowing steering to be disabled. Since every action is decoded by the frozen $\pidpw$, steering changes which actions are selected without modifying the policy itself. When trained in imagination, however, steering can exploit prediction errors in the learned world model. Following DiWA \citep{chandra2025diwa}, we add a behavior-cloning term $\mathcal{L}_{\mathrm{BC}}$ that keeps decoded actions close to those produced from prior noise (Appendix, Eq.~\ref{eq:app_bc}).\looseness=-1 

\subsection{Rehearsal in the World Model}
\label{sec:rehearsal}

Training the steering module on the current task can change how it steers other tasks, causing forgetting. To preserve prior skills, we relax the hard retention constraint in Eq.~\ref{eq:objective} into a soft penalty between the current steering policy and a task-specific reference. For each task $\ell_j \in \mathcal{R}_k$, the reference $\smash{\piw_{\mathrm{ref}(j)}}$ is $\smash{\piw_{k-1}}$ if the task was previously steered, and the prior $\mathcal{N}(0,I)$ otherwise. Thus, pretrained skills are anchored to the unsteered policy, while previously steered skills are anchored to their earlier steering behavior. Since the current steering policy and its reference are both diagonal Gaussians, their KL divergence has a closed form. We average it over the retention set,
\begin{equation}
\label{eq:kl}
\mathcal{L}_{\mathrm{KL}}(\piw_k) = \frac{1}{|\mathcal{R}_k|} \sum_{\ell_j \in \mathcal{R}_k} \mathbb{E}_{s \sim \mathcal{D}^{j}_{\mathrm{reh}}}\Big[\mathrm{KL}\big(\piw_k(\cdot \mid s,\ell_j)\,\big\|\,\piw_{\mathrm{ref}(j)}(\cdot \mid s,\ell_j)\big)\Big].
\end{equation}

Here, $\smash{\mathcal{D}^{j}_{\mathrm{reh}}}$ contains states of task $\ell_j$ generated by the world model for rehearsal. For each task $\ell_j \in \mathcal{R}_k$, a rollout is seeded from one of the task's initial observations, encoded by $E_\phi$ , and unrolled with $P^\cW_\phi$, while the reference policy $\smash{\piw_{\mathrm{ref}(j)}}$ selects the noise at each step. Rehearsal requires no reward, and rollouts run for a fixed horizon unless a success signal is available for early termination. Rollouts are regenerated at every update rather than stored, requiring world model compute but no trajectory buffer. Since the world model is frozen, the rehearsal distribution does not drift as the number of tasks grows. Rollouts follow the reference policy, so the KL penalty of Eq.~\ref{eq:kl} anchors the current module on states relevant to the retained skill.

\subsection{World Model as Simulator or Memory}
\label{sec:two_roles}

D2R uses the world model either as a simulator or as memory, and the two roles require different fidelity. As a simulator, the model must be accurate because the steering policy is optimized against it and can exploit prediction errors. As memory, it only needs to generate states representative of retained tasks, since the KL penalty matches a reference on those states rather than optimizing a policy through the world model dynamics. Whether the model serves as simulator or memory therefore depends on its coverage of the task. A model trained on data that covers the task dynamics can support optimization, while one outside that coverage may still serve as memory, so the same world model can play different roles across tasks. This assignment determines where the current task is learned, in the imagined latent-noise MDP $\Mwm^{\cW}$ when the model serves as a simulator and in the environment latent-noise MDP $\Menv^{\cW}$ when it serves only as memory. Both regimes admit any RL algorithm over $\cW$, but their interaction costs differ. In $\Mwm^{\cW}$, rollouts are cheap and parallelizable, so we use on-policy PPO with abundant on-policy data \citep{schulman2017ppo,chandra2025diwa}. In $\Menv^{\cW}$, every transition is a real episode, so we use off-policy  SAC and reuse real experience through replay \citep{haarnoja2018sac,nematollahi2022robot,wagenmaker2025dsrl}. Rehearsal is the same in both regimes, since retained tasks are always rehearsed in the world model and Eq.~\ref{eq:kl} needs no environment interaction.

At stage $k$, we train a new module $\smash{\piw_k}$ from scratch with $\smash{\piw_{k-1}}$ frozen as the reference, minimizing
\begin{equation}
\label{eq:total}
\mathcal{L}(\piw_k) = \mathcal{L}_{\mathrm{RL}} + \alpha\, \mathcal{L}_{\mathrm{BC}} + \lambda\, \mathcal{L}_{\mathrm{KL}},
\end{equation}
where $\alpha$ is the behavior-cloning coefficient, set to $0$ for training with environment interaction, and $\lambda$ is the retention coefficient shared across retained tasks. After stage $k$, $\smash{\piw_k}$ becomes the sole steering module for all tasks seen so far, and $\smash{\piw_{k-1}}$ is discarded (See Algorithm~\ref{alg:d2r} in the Appendix).

%%%%%%%%%%%%%%%%%%%%%%%%%%%%%%%%%%%%%%%%%%%%%%%%%%%%%%%%%%%%%%%%%%%%%%%%%%%%%%%%
\section{Simulation Experiments}
\label{sec:sim_experiments}

We evaluate three aspects of D2R in simulation, namely steering in imagination for improving a task without forgetting others (\S\ref{sec:sim_single}), continual steering with world model rehearsal for learning tasks sequentially while retaining earlier ones (\S\ref{sec:sim_continual}), and rehearsal with a coarse world model (\S\ref{sec:sim_memory}).

\subsection{Experimental Setup}
\label{sec:sim_setup}
We evaluate D2R on two benchmarks spanning its two regimes, where the world model serves as both simulator and memory or as memory alone (\S\ref{sec:two_roles}). We follow two protocols. In \emph{single-task adaptation}, a method is trained on one task and evaluated on all tasks, measuring improvement on adapted task and retention of the others. In \emph{sequential adaptation}, a model learns tasks sequentially and is evaluated on all tasks after each stage, measuring forgetting and retention over the sequence.\looseness=-1 

\textbf{CALVIN.}~On CALVIN \citep{mees2022calvin}, adaptation occurs entirely in imagination. The world model is a recurrent state-space model (RSSM) trained on task-agnostic play data from environment, while the frozen policy is a language-conditioned multi-task diffusion policy trained on 150 demonstrations per task. Since imagined rollouts are cheap and parallel, each task is improved with on-policy PPO in the world model, rewarded by a latent success classifier, and eight tasks are adapted in isolation and in sequence, rehearsing the other seven at each adaptation (see Appendices~\ref{sec:app_impl} and~\ref{sec:app_setup} for implementation details and the adaptation order).  

\textbf{LIBERO.}~On LIBERO \citep{liu2023libero}, we use an off-the-shelf policy and world model. The policy is the public $\pi_0$ checkpoint \citep{black2024pi_0}, fine-tuned on the Spatial, Object, Goal, and Long suites, while the world model is A2World \citep{huang2026learning}, pretrained on 2.1M real-robot trajectories and fine-tuned on the same suites, with neither exposed to LIBERO-90. We learn eight LIBERO-90 tasks sequentially (order in Appendix~\ref{sec:app_setup}). Since A2World is not accurate enough for policy optimization on these tasks (see \S\ref{sec:sim_memory}), each is learned in the environment, where every transition requires real interaction. All methods therefore learn off-policy with SAC and replay under the same interaction budget, while A2World serves only for rehearsal. Each stage rehearses the previously learned LIBERO-90 tasks and the four suites in A2World with reward-free rollouts.\looseness=-1 

\textbf{D2R.}~In both benchmarks, the steering module is trained from scratch at each stage. To seed rehearsal rollouts, we store only the first observation of each demonstration, giving 150 observations per task on CALVIN and 50 on LIBERO (see Appendix~\ref{sec:app_hparams} for hyperparameters).

\textbf{Baselines.}~We compare against the frozen base policy and three families of alternatives, corresponding to the three spaces in which adaptation can occur. Weight-space adaptation is represented by \textbf{DiWA} \citep{chandra2025diwa}, which fine-tunes all policy weights in imagination, in single-task adaptation, and by \textbf{LoRA} fine-tuning \citep{hu2022lora} in sequential adaptation. To separate the effect of rehearsal from the adaptation space, \textbf{DiWA + rehearsal} augments DiWA with our world model rehearsal, regularizing the fine-tuned policy toward the frozen base policy on imagined rehearsal states. LoRA trains low-rank adapters, on the full policy network on CALVIN and on the action expert of $\pi_0$ on LIBERO, with PPO in imagination on CALVIN and off-policy learning in the environment on LIBERO. Output-space adaptation is represented by \textbf{DP-TTC}, a test-time selection method on CALVIN in the style of \citet{cui2026dreamsteer}. It samples action chunks from the frozen policy, imagines their outcomes in the world model, scores them with the success classifier, and executes one of the highest-scoring candidates. Input-space adaptation without rehearsal is represented by \textbf{DSRL} \citep{wagenmaker2025dsrl}, trained in the world model on CALVIN and in the environment on LIBERO. We also consider three ablations that add the rehearsal penalty to DSRL and differ only in the states it is applied to. \textbf{DSRL + instruction KL} applies the rehearsal penalty to current-task observations conditioned on retained-task instructions, without storing or generating states from the retained tasks. \textbf{DSRL + first-obs KL} applies it only at the stored first observations, without rolling them forward through the world model. \textbf{DSRL + replay KL} applies the penalty to retained-task rollouts collected once per stage with the previous steering module and stored (real trajectories on LIBERO and imagined trajectories on CALVIN). \textbf{D2R} instead applies the penalty to retained-task rollouts regenerated in the world model. All methods start from the same base policy and use the same world model and seed frames where applicable (see Appendix~\ref{sec:app_baselines} for baseline details).

\textbf{Metrics.}~We report three quantities. The first is the \textbf{success rate}. On CALVIN, results average success over all validation initial states per task and three training seeds. On LIBERO, they average 30 evaluation episodes from the benchmark's initial states over three evaluation seeds for one model. We report 95\% confidence intervals throughout. We define \textbf{retention} as the change in success, in percentage points, relative to the frozen base policy. In sequential adaptation, we define \textbf{forgetting} as the drop relative to the stage at which the task was learned (Appendix~\ref{sec:app_metrics}). Retention measures performance relative to the original policy, while forgetting measures only the loss after learning.\looseness=-1 
\begin{table*}[t]
    \centering
    \caption{
    \textbf{Single-task adaptation on CALVIN.} Success on the adapted task and retention, defined as the mean change in success over all tasks relative to the base policy in percentage points (\upglyph\ gain, \dnglyph\ loss). Baselines trade retention for task gains, while D2R is the only method to improve both.
    }
    \label{tab:calvin_results_v2}
    % One line per cell: success rate +- 95% CI on the adapted task (as in Table 1), followed in
% parentheses by the mean change on the other seven tasks relative to the base policy.
%   red triangle   : drop of more than 2 points
%   gray tilde     : within +-2 points (no change)
%   green triangle : gain of more than 2 points
% Base DP and DP-TTC are not adapted, so they carry no parenthesis.
% Deltas of 0.0 are placeholders until the cross-task numbers are in.
% Triangles are drawn with tikz.

\providecommand{\retthresh}{0.05} % changes that round to 0.0 get no triangle
\providecommand{\dnmark}[1]{\textcolor{red!80!black}{\tikz[baseline=0pt]\fill (0,0.95ex)--(0.9ex,0.95ex)--(0.45ex,0.1ex)--cycle;\,$#1$}}
\providecommand{\upmark}[1]{\textcolor{green!55!black}{\tikz[baseline=0pt]\fill (0,0.1ex)--(0.9ex,0.1ex)--(0.45ex,0.95ex)--cycle;\,$#1$}}
\providecommand{\upglyph}{\textcolor{green!55!black}{\tikz[baseline=0pt]\fill (0,0.1ex)--(0.9ex,0.1ex)--(0.45ex,0.95ex)--cycle;}}
\providecommand{\dnglyph}{\textcolor{red!80!black}{\tikz[baseline=0pt]\fill (0,0.95ex)--(0.9ex,0.95ex)--(0.45ex,0.1ex)--cycle;}}
\providecommand{\nomark}[1]{\textcolor{gray}{$#1$}}
\providecommand{\retdelta}[1]{%
  \pgfmathsetmacro{\absd}{abs(#1)}%
  \ifdim #1pt<-\retthresh pt \dnmark{\pgfmathprintnumber[fixed,fixed zerofill,precision=1]{\absd}}%
  \else\ifdim #1pt>\retthresh pt \upmark{\pgfmathprintnumber[fixed,fixed zerofill,precision=1]{\absd}}%
  \else \nomark{\pgfmathprintnumber[fixed,fixed zerofill,precision=1]{\absd}}\fi\fi}
% adapted-task success, 95% CI half-width, signed delta on other tasks
\providecommand{\cellstat}[3]{$#1 \pm #2$\,(\retdelta{#3})}
% split layout: success and retention as separate sub-columns
\providecommand{\onedec}[1]{\pgfmathprintnumber[fixed,fixed zerofill,precision=1]{#1}}
\providecommand{\scell}[2]{$\onedec{#1}\mkern1.5mu{\pm}\mkern1.5mu\onedec{#2}$}
\providecommand{\scellB}[2]{{\bfseries\boldmath$\onedec{#1}\mkern1.5mu{\pm}\mkern1.5mu\onedec{#2}$}}
\providecommand{\rcell}[1]{\retdelta{#1}}
\providecommand{\rcellB}[1]{{\boldmath\retdelta{#1}}}
% bold variants: best success in the row (B), best retention in the row (R)
\providecommand{\cellstatB}[3]{{\boldmath$#1 \pm #2$}\,(\retdelta{#3})}
\providecommand{\cellstatR}[3]{$#1 \pm #2$\,({\boldmath\retdelta{#3}})}
% nothing yet
\providecommand{\cellna}{--\,\textcolor{gray}{(--)}}

\renewcommand{\arraystretch}{1.15}
\setlength{\tabcolsep}{2pt}

{\fontsize{6}{7.2}\selectfont% 8pt body, no resizebox
\begin{tabular*}{\linewidth}{@{\extracolsep{\fill}}c|l|c|c|cc|cc|cc|cc@{}}
    \toprule
    \multirow{2}{*}[-3pt]{\textbf{ID}} & \multirow{2}{*}[-3pt]{\textbf{Task}} & \textbf{Base DP} & \textbf{DP-TTC}
    & \multicolumn{2}{c|}{\textbf{DiWA}} & \multicolumn{2}{c|}{\textbf{DiWA + rehearsal}} & \multicolumn{2}{c|}{\textbf{DSRL}} & \multicolumn{2}{c}{\textbf{D2R (Ours)}} \\
    \cmidrule(lr){3-3}\cmidrule(lr){4-4}\cmidrule(lr){5-6}\cmidrule(lr){7-8}\cmidrule(lr){9-10}\cmidrule(lr){11-12}
    & & Success & Success & Success & Retention & Success & Retention & Success & Retention & Success & Retention \\
    \midrule
    1 & close-drawer & \scell{77.01}{4.95} & \scell{79.31}{8.57} & \scell{88.51}{13.08} & \rcell{-41.68} & \scell{94.25}{4.97} & \rcell{-11.36} & \scellB{98.85}{4.95} & \rcell{-3.73} & \scell{97.7}{4.95} & \rcellB{2.23}  \\
        2 & move-slider-left & \scell{39.59}{16.16} & \scell{43.75}{13.45} & \scell{48.96}{4.47} & \rcell{-52.89} & \scell{59.37}{3.01} & \rcell{-12.76} & \scellB{62.50}{15.53} & \rcell{-5.07} & \scell{54.17}{33.84} & \rcellB{0.02}  \\
    3 & move-slider-right & \scell{42.53}{32.42} & \scell{49.42}{9.88} & \scell{68.97}{8.56} & \rcell{-48.15} & \scell{74.71}{4.95} & \rcell{-16.8} & \scellB{85.06}{9.90} & \rcell{-0.77} & \scell{79.31}{13.09} & \rcellB{3.72}  \\
    4 & open-drawer & \scell{50.00}{8.27} & \scell{41.11}{4.78} & \scellB{87.78}{12.65} & \rcell{-46.88} & \scell{63.33}{21.91} & \rcell{-21.75} & \scell{78.90}{4.73} & \rcell{-3.07} & \scell{77.78}{9.57} & \rcellB{3.58}  \\
    5 & turn-off-LED & \scell{85.41}{16.16} & \scell{87.50}{13.44} & \scellB{96.88}{7.76} & \rcell{-35.91} & \scell{94.79}{8.97} & \rcell{-10.41} & \scell{91.66}{4.49} & \rcell{-6.97} & \scell{86.46}{8.95} & \rcellB{0.53}  \\
    6 & turn-off-lightbulb & \scell{40.23}{26.17} & \scell{43.68}{9.90} & \scellB{60.92}{13.09} & \rcell{-43.39} & \scell{51.72}{12.12} & \rcell{-10.16} & \scell{48.28}{8.56} & \rcell{-5.49} & \scell{58.62}{21.55} & \rcellB{0.55}  \\
    7 & turn-on-LED & \scell{82.76}{14.84} & \scell{85.06}{4.95} & \scell{90.81}{13.08} & \rcell{-35.15} & \scellB{96.55}{8.57} & \rcell{-10.01} & \scell{91.95}{13.08} & \rcell{-10.48} & \scell{88.51}{9.88} & \rcellB{1.23}  \\
    8 & turn-on-lightbulb & \scell{91.92}{4.35} & \scell{90.91}{13.04} & \scell{90.91}{7.53} & \rcell{-47.47} & \scell{88.89}{19.92} & \rcell{-8.14} & \scellB{96.97}{7.53} & \rcell{-6.95} & \scell{95.96}{4.35} & \rcellB{1.07}  \\
    \midrule
    & \textbf{Average} & \scell{63.68}{6.52} & \scell{65.09}{1.93} & \scell{79.22}{3.66} & \rcell{-43.94} & \scell{77.95}{3.10} & \rcell{-12.67} & \scellB{81.77}{2.86} & \rcell{-5.34} & \scell{79.81}{3.79} & \rcellB{1.62}  \\
    \bottomrule
\end{tabular*}%
}

\end{table*}
\subsection{Steering in Imagination Without Forgetting}
\label{sec:sim_single}
Table~\ref{tab:calvin_results_v2} reports single-task adaptation on the eight CALVIN tasks, with all methods trained entirely in the world model and no environment interaction. D2R improves the frozen policy on every task, raising average success from 63.7\% (Base DP) to 79.8\%, and is the only method with positive retention. Average retention is $-43.9$ points for DiWA, $-12.7$ for DiWA with rehearsal, $-5.3$ for DSRL, and $+1.6$ for D2R. In success rate, D2R matches DiWA, which fine-tunes the policy weights in imagination, showing that input-noise steering can achieve the same gains without updating the policy. DSRL, which steers without rehearsal, scores two points higher on the target task, while D2R trades this small margin for retaining prior skills. DP-TTC leaves the policy unchanged and thus induces no forgetting, but underperforms the adaptation methods despite querying the world model and classifier at every step, as it can only select among actions proposed by the frozen policy.

These results reflect the consequences of the adaptation space. Fine-tuning updates the shared policy weights, so learning one task changes the actions produced on all others, and DiWA loses retention regardless of which task is adapted. Steering leaves the policy frozen and changes only which noise is fed to it, so every action is still produced by the same pretrained policy \citep{yu2026score,murray2026flowdagger}, while the rehearsal penalty constrains the module on the other tasks. Adding the same rehearsal to DiWA recovers most, but not all, of the retention loss. Rehearsal constrains the fine-tuned policy only on imagined states visited by the reference, whereas changed weights can still alter behavior elsewhere. In the noise space, an unsteered task also has a simple reference, the Gaussian prior. Rehearsal thus benefits both adaptation spaces, but is most effective when combined with steering (see Figure~\ref{fig:app_single_heatmaps} for retention heatmaps). 

\vspace{-0.1cm}
\subsection{Continual Steering Across Tasks}
\label{sec:sim_continual}

\begin{wrapfigure}[27]{r}{0.28\textwidth}
    \vspace{-72pt}
    \centering
    % Three-panel vertical stack (DSRL, DSRL + instruction KL, D2R) for a narrow wrapfigure. Sizes in cm on purpose (pgfplots mis-sizes \textwidth inside boxes).
% LIBERO sequence forgetting-score heatmaps (DSRL + instruction KL, D2R) for the wrapped Figure 3.
% Cells: diagonal = success at learning (blue), below = points lost since learned (red), blank above.
% Same styles and data as results/calvin_retention_heatmaps_sample.tex.
\begingroup
\input{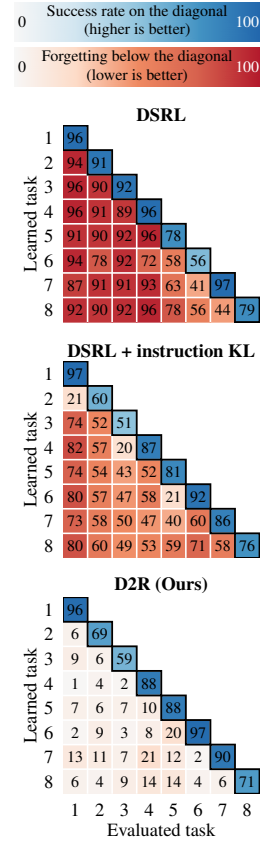}
\pgfplotsset{
  seqheat/.style={
    width=2.6cm, height=2.6cm, scale only axis, axis equal image,
    xmin=-0.5, xmax=7.5, ymin=-0.5, ymax=7.5, y dir=reverse, enlargelimits=false, axis on top,
    tick style={draw=none}, xtick={0,...,7}, ytick={0,...,7},
    xticklabels={1,2,3,4,5,6,7,8}, yticklabels={1,2,3,4,5,6,7,8},
    x tick label style={font=\scriptsize}, y tick label style={font=\scriptsize},
    title style={font=\scriptsize\bfseries, at={(0.5,1)}, anchor=south, yshift=-8pt},
    label style={font=\scriptsize}, xlabel shift=-4pt, ylabel shift=-4pt,
    colormap={learnforg}{rgb255=(33,102,172) rgb255=(67,147,195) rgb255=(146,197,222) rgb255=(209,229,240) rgb255=(247,247,247) rgb255=(253,219,199) rgb255=(239,138,98) rgb255=(214,96,77) rgb255=(178,24,43)},
    point meta min=-100, point meta max=100, axis line style={draw=none},
  },
  seqcells/.style={
    matrix plot*, mesh/cols=8, point meta=explicit,
    visualization depends on={\thisrow{succ} \as \succval},
    nodes near coords={\ifdim\succval pt<-900pt\else\pgfmathprintnumber[fixed,precision=0]{\succval}\fi},
    nodes near coords style={font=\tiny, anchor=center, /pgf/number format/assume math mode=true,
      /utils/exec={\pgfmathfloattofixed{\pgfplotspointmeta}\let\metafixed\pgfmathresult\pgfmathparse{abs(\metafixed)>55 ? 1 : 0}\ifnum\pgfmathresult=1 \color{white}\else\color{black}\fi}},
  },
}
\newcommand{\seqgrid}{\pgfplotsinvokeforeach{0.5,1.5,2.5,3.5,4.5,5.5,6.5}{\draw[white,line width=0.6pt] (axis cs:##1,-0.5)--(axis cs:##1,7.5); \draw[white,line width=0.6pt] (axis cs:-0.5,##1)--(axis cs:7.5,##1);}}
\newcommand{\seqblank}{\pgfplotsinvokeforeach{0,1,2,3,4,5,6}{\fill[white] (axis cs:##1+0.5,##1-0.5) rectangle (axis cs:7.5,##1+0.5);}}
\newcommand{\seqdiag}{\pgfplotsinvokeforeach{0,1,2,3,4,5,6,7}{\draw[black,line width=0.7pt] (axis cs:##1-0.5,##1-0.5) rectangle (axis cs:##1+0.5,##1+0.5);}}
\begin{tikzpicture}
\begin{groupplot}[group style={group size=1 by 3, vertical sep=0.52cm}, seqheat]
\nextgroupplot[title={DSRL}, ylabel={Learned task}, xticklabels={}]
\addplot[seqcells] table[x=x,y=y,meta=d] {\heatDsrlSeqLiberoFs};
\seqblank\seqgrid\seqdiag
\nextgroupplot[title={DSRL + instruction KL}, ylabel={Learned task}, xticklabels={}]
\addplot[seqcells] table[x=x,y=y,meta=d] {\heatDtrNoMemSeqLiberoFs};
\seqblank\seqgrid\seqdiag
\nextgroupplot[title={D2R (Ours)}, ylabel={Learned task}, xlabel={Evaluated task}]
\addplot[seqcells] table[x=x,y=y,meta=d] {\heatDtrSeqLiberoFs};
\seqblank\seqgrid\seqdiag
\end{groupplot}
\path (group c1r1.outer north west) coordinate (cbL) (group c1r1.north east) coordinate (cbR);
\pgfpointdiff{\pgfpointanchor{cbL}{center}}{\pgfpointanchor{cbR}{center}}\pgfgetlastxy{\cbW}{\cbH}
\pgfkeys{/pgfplots/cbwidth/.initial=\cbW}
\begin{axis}[at={(cbL)}, anchor=south west, yshift=5pt, width={\pgfkeysvalueof{/pgfplots/cbwidth}}, height=0.52cm, scale only axis,
  hide axis, xmin=0, xmax=100, ymin=0, ymax=1, point meta min=0, point meta max=100,
  colormap={forg}{rgb255=(247,247,247) rgb255=(253,219,199) rgb255=(239,138,98) rgb255=(214,96,77) rgb255=(178,24,43)}]
\addplot[surf, shader=interp, point meta=x, mesh/cols=5, mesh/rows=2] coordinates {(0,0) (25,0) (50,0) (75,0) (100,0) (0,1) (25,1) (50,1) (75,1) (100,1)};
\node[font=\tiny] at (axis cs:3,0.5) {0};
\node[font=\tiny, align=center] at (axis cs:50,0.5) {Forgetting below the diagonal\\(lower is better)};
\node[font=\tiny, white] at (axis cs:95,0.5) {100};
\end{axis}
\begin{axis}[at={(cbL)}, anchor=south west, yshift=22pt, width={\pgfkeysvalueof{/pgfplots/cbwidth}}, height=0.52cm, scale only axis,
  hide axis, xmin=0, xmax=100, ymin=0, ymax=1, point meta min=0, point meta max=100,
  colormap={learn}{rgb255=(247,247,247) rgb255=(209,229,240) rgb255=(146,197,222) rgb255=(67,147,195) rgb255=(33,102,172)}]
\addplot[surf, shader=interp, point meta=x, mesh/cols=5, mesh/rows=2] coordinates {(0,0) (25,0) (50,0) (75,0) (100,0) (0,1) (25,1) (50,1) (75,1) (100,1)};
\node[font=\tiny] at (axis cs:3,0.5) {0};
\node[font=\tiny, align=center] at (axis cs:50,0.5) {Success rate on the diagonal\\(higher is better)};
\node[font=\tiny, white] at (axis cs:95,0.5) {100};
\end{axis}
\end{tikzpicture}
\endgroup
    \vspace{-0.2cm}
    \caption{\textbf{Continual steering on LIBERO.} Diagonal cells show the success when each task is learned, and cells below show the forgetting, the points lost after each later stage. DSRL forgets each task as soon as the next is learned, instruction KL reduces this, and D2R keeps the tasks to the end.}    \label{fig:calvin_retention_heatmaps}\label{fig:calvin_sequence}
    \vspace{-10pt}
\end{wrapfigure}

\vspace{-0.1cm}
We next ask whether a single steering module can learn tasks one after another without forgetting. The two benchmarks test this in different ways. On CALVIN, the frozen policy already performs all eight tasks in a single scene, and the module refines them one at a time in imagination. On LIBERO, all eight tasks are new to the policy and span different scenes. Since the world model is not accurate for policy optimization, each task is learned in the environment while the world model is used only for rehearsal. Table~\ref{tab:sequence} reports the average \textbf{success} over all tasks at the end of the sequence, together with \textbf{retention} and \textbf{forgetting} as defined above (see Table~\ref{tab:app_sequence_full} for per-stage columns). Figure~\ref{fig:calvin_sequence} shows the LIBERO sequence stage by stage, where each row is a stage, the outlined diagonal cell gives the success on the task learned at that stage, and the cells below give how many points that task has lost since. D2R learns each sequence with a single module, raising average success from 63.7\% to 74.8\% on CALVIN and from 10.4\% to 74.9\% on LIBERO, with only 5.7 and 8.3 points of forgetting, while the four LIBERO suites remain at their base performance ($+1.6$ retention).
\noindent\begin{minipage}{0.69\textwidth}
    \captionof{table}{Continual steering over eight tasks on CALVIN (in imagination) and LIBERO (in the environment). Storage is what each method keeps besides the frozen policy after the full sequence.}
    \label{tab:sequence}
    % Compact Table 2 for the main paper: Success, Retention, Forgetting per benchmark, with Storage inside the LIBERO block.
% Full version with Single and New: results/sequence_results.tex (appendix).
\setlength{\tabcolsep}{0.9pt}
\renewcommand{\arraystretch}{1.05}
{\fontsize{6}{7.2}\selectfont
\begin{tabular*}{\linewidth}{@{\extracolsep{\fill}}lccc|cccc@{}}
\toprule
& \multicolumn{3}{c|}{\textbf{CALVIN}} & \multicolumn{4}{c}{\textbf{LIBERO}} \\
\cmidrule(lr){2-4}\cmidrule(lr){5-8}
\textbf{Method} & Success$\uparrow$ & Retention$\uparrow$ & Forgetting$\downarrow$ & Success$\uparrow$ & Retention$\uparrow$ & Forgetting$\downarrow$ & Storage$\downarrow$ \\
\midrule
    Base & $63.7 \pm 6.5$ & $0.0$ & $0.0$ & $10.4 \pm 2.7$ & $0.0$ & $0.0$ & -- \\
    LoRA & $11.7 \pm 1.9$ & $-52.0$ & $53.2$ & $1.0\pm 1.6$ & $-41.6$ & $61.4$ & -- \\
    DSRL & $63.6 \pm 4.8$ & $-0.1$ & $\mathbf{1.6}$ & $16.9\pm 5.2$ & $-30.8$ & 78.3 & -- \\
    \hspace{0.5em}+ instruction KL & $70.9 \pm 8.1$ & $+7.2$ & 3.6 & $24.9\pm 2.6$ & $-23.3$ & 61.4 & -- \\
    \hspace{0.5em}+ first-obs KL & $65.1 \pm 3.6$ & $+1.4$ & 12.9 & $36.4 \pm 0.6$ & $-20.9$ & 52.5 & $\sim$400 frames \\
    \hspace{0.5em}+ replay KL & $74.0 \pm 3.5$ & $+10.3$ & 5.8 & $73.9\pm 6.3$ & $\mathbf{+2.0}$ & 11.6 & $\sim$59k frames \\
    \textbf{D2R (Ours)} & {\boldmath$74.8 \pm 3.7$} & $\mathbf{+11.1}$ & 5.7 & {\boldmath$74.9\pm 2.6$} & $+1.6$ & $\mathbf{8.3}$ & $\sim$400 frames \\
\midrule
    \textit{Per-task modules} & $81.8 \pm 2.9$ & $+18.1$ & $0.0$ & $91.3 \pm 1.0$ & $+80.9$ & $0.0$ & 8 modules \\
    \textit{Multi-task module} & $67.3 \pm 3.9$ & $\times$ & $\times$  & $54.6 \pm 8.5$ & $\times$ & $\times$ & 1 module \\
\bottomrule
\end{tabular*}}

\end{minipage}
\vspace{6pt}

Neither weight updates nor steering without rehearsal achieves this balance of adaptation and retention. LoRA fine-tuning learns each task but quickly forgets earlier ones, losing 53.2 points on CALVIN and 61.4 on LIBERO, and finishes well below the base policy on both benchmarks. DSRL, which steers without rehearsal, also ends near the base level on both, but for different reasons. On LIBERO, it forgets 78.3 points and reduces performance on the four pretrained suites by 30.8 points. The DSRL panel of Figure~\ref{fig:calvin_sequence} similarly shows each LIBERO task turning red one stage after it is learned. On CALVIN, however, DSRL forgets only 1.6 points because it never improves much beyond the base policy. Without rehearsal, adaptation therefore either overwrites earlier skills or fails to accumulate gains across tasks. The three ablations differ from D2R only in the states used for rehearsal, revealing what the memory must provide. DSRL + instruction KL uses no states from retained tasks and reaches 70.9\% on CALVIN but only 24.9\% on LIBERO, a gap we explain in \S\ref{sec:sim_memory}. DSRL + first-obs KL uses only the stored first observations, and forgets 12.9 points on CALVIN and 52.5 on LIBERO. DSRL + replay KL uses full rollouts of the retained tasks, collected once per stage and stored, and comes closest. D2R achieves comparable performance while keeping about 400 frames over the whole sequence, roughly 150 times fewer than a trajectory buffer. Rehearsal therefore requires states that retained skills actually visit, but those states need not be real or stored.
We also compare against two reference settings. Task-specific modules, one per task with the task identity given at test time, reach 81.8\% on CALVIN and 91.3\% on LIBERO and cannot forget by construction. D2R comes within 7 and 16 points of them using a single module. A single module trained jointly on all tasks reaches only 67.3\% and 54.6\% despite a larger budget, because joint training must optimize sparse rewards for all eight tasks simultaneously, whereas rehearsal anchors each retained task to a reference that already solves it. D2R keeps one module and a few initial frames per task, and among methods that neither grow with the sequence nor store trajectories, it is the only one that both learns the tasks and retains them.

\subsection{A Coarse World Model as Memory}
\label{sec:sim_memory}

On LIBERO the world model serves only as memory, because it is not accurate enough to learn in. A2World was fine-tuned on the four LIBERO suites but never saw LIBERO-90. To test whether it can serve as a simulator for the target tasks, we clone real demonstrations of each task and A2World-imagined replays of the same demonstrations with the same behavior-cloning recipe (Appendix~\ref{sec:app_faithfulness_check}). Policies cloned from real demonstrations reach 74.9\% on the LIBERO-90 tasks and those cloned from imagined replays only 19.6\%, whereas on the four suites the two are within a few points. The model reproduces the dynamics it was trained on but not the target tasks, so a steering policy optimized inside it would exploit its prediction errors rather than learn the task. The same model nonetheless serves as memory. The three panels of Figure~\ref{fig:calvin_sequence} differ only in what anchors the module, with darker red indicating stronger forgetting. Without rehearsal, every column of DSRL turns dark red within one stage. With instruction KL the red is lighter, and with D2R the columns remain pale to the end. The difference is where the penalty is applied. Instruction KL anchors the module on observations from the current task paired with retained instructions. This works on CALVIN, where all tasks share a scene, but not on LIBERO, where scenes change and current-task observations do not cover retained-task states. D2R obtains these states by unrolling each retained task in A2World from stored initial frames. Although these imagined trajectories are too inaccurate to learn from, their states are sufficient for anchoring, and rehearsal on them matches rehearsal on real stored trajectories. A model accurate enough to optimize against is therefore a stronger requirement than one accurate enough to rehearse in. An off-the-shelf world model can serve as memory without any fine-tuning on the tasks being learned, for the pretrained suites and the new tasks alike.

%%%%%%%%%%%%%%%%%%%%%%%%%%%%%%%%%%%%%%%%%%%%%%%%%%%%%%%%%%%%%%%%%%%%%%%%%%%%%%%%
\section{Real-World Experiments}
\label{sec:real_experiments}

\begin{figure*}[t]
    \centering
    \input{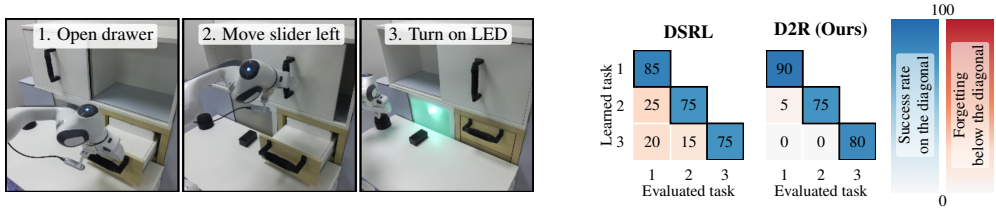}
    \vspace{-0.2cm}
    \caption{\textbf{Continual steering on a real Franka robot.} Heatmaps follow Figure~\ref{fig:calvin_sequence}, with the success rate when each task is learned on the diagonal and the forgetting score below it. D2R retains prior gains across the sequence, whereas DSRL without rehearsal loses about half of them. Adaptation occurs entirely in imagination, with the robot used only for evaluation.}
    \vspace{-0.3cm}
    \label{fig:franka}
\end{figure*}

\textbf{Setup.}~We evaluate D2R on a Franka Emika Panda robot in a tabletop environment with a cabinet and drawer, observed by a static and a gripper-mounted camera. The world model is an RSSM trained once on six hours of teleoperated play data and then frozen. The base policy is a single language-conditioned diffusion policy trained on 100 demonstrations for each of seven tabletop tasks (see Appendix~\ref{sec:app_real_setup}), with one success classifier for each task trained on the same demonstrations. The policy, classifiers, and world model remain frozen throughout. We adapt three of these tasks sequentially, open drawer, move slider left, and turn on LED, rehearsing prior tasks at each later stage. As on CALVIN, the world model is trained on the robot's own play data and is sufficiently faithful for policy optimization, so learning and rehearsal take place in imagination with PPO, and the robot is used only for evaluation. After each stage, we run 20 episodes per adapted task from fixed initial configurations and report success rates. We compare against the frozen base policy and DSRL trained in the same world model without rehearsal (see Appendix~\ref{sec:app_hparams} for hyperparameters).

\textbf{Results.}~Figure~\ref{fig:franka} shows the three adaptation stages. For each adapted task, steering in imagination improves execution on the real robot by 40, 20, and 30 points over a frozen base policy that initially succeeds on 50 to 55\% of episodes. The methods diverge only after later tasks are learned. Under D2R, those gains are preserved to the end of the sequence, for an average of 30 points above the base policy. Under DSRL, without rehearsal, about half of them are lost, leaving an average of 15 points. World model rehearsal thus preserves prior gains on the real robot without robot time for learning.\looseness=-1 

%%%%%%%%%%%%%%%%%%%%%%%%%%%%%%%%%%%%%%%%%%%%%%%%%%%%%%%%%%%%%%%%%%%%%%%%%%%%%%%%
\section{Conclusion}

We presented D2R, a framework for adapting a pretrained generative robot policy to tasks arriving sequentially without forgetting its existing skills. D2R steers the frozen policy in its noise space with a single small module and rehearses the retained skills on rollouts imagined in a frozen world model. Adaptation therefore requires no weight updates, no stored trajectories, and no robot interaction for rehearsal. On two simulated benchmarks and a real Franka robot, steering matches weight fine-tuning on the adapted task while preserving prior skills substantially better. Our experiments further show that a world model needs lower fidelity for rehearsal than for policy optimization, so a model too inaccurate to learn in can still serve as memory.
However, D2R still stores a few initial frames per task to prompt the world model, which future work could instead generate on demand.\looseness=-1
\subsection*{Acknowledgements}
This work was partially funded by the Carl Zeiss Foundation with the ReScaLe project.
\bibliography{iclr2027_conference}
\bibliographystyle{iclr2027_conference}

\appendix
\setcounter{section}{0}
\setcounter{equation}{0}
\setcounter{table}{0}
\setcounter{figure}{0}
\renewcommand{\thesection}{S.\arabic{section}}
\renewcommand{\thesubsection}{S.\arabic{section}.\arabic{subsection}}
\renewcommand{\theequation}{S.\arabic{equation}}
\renewcommand{\thetable}{S.\arabic{table}}
\renewcommand{\thefigure}{S.\arabic{figure}}
\renewcommand{\thealgorithm}{S.\arabic{algorithm}}

% Table of contents for the supplementary material only.
\etocdepthtag.toc{appendix}
\etocsettagdepth{mtappendix}{none}
\etocsettagdepth{appendix}{subsection}
\etocsetstyle{section}{\vspace{0.6em}}{}{\noindent\textbf{\etocnumber\hspace{0.8em}\etocname}\hfill\textbf{\etocthepage}\par}{}
\etocsetstyle{subsection}{}{}{\noindent\hspace{1.8em}\etocnumber\hspace{0.8em}\etocname\dotfill\etocthepage\par\vspace{0.15em}}{}
\etocsettocstyle{\section*{Appendix}}{}
\clearpage
\tableofcontents
\clearpage

\section{Training Objectives}
\label{sec:app_objectives}
This section states all objectives used by D2R, in the order they appear in Eq.~\ref{eq:total}. Throughout, $s$ denotes the state representation conditioned on by the steering module, $\ell$ an instruction, and $\bw \in \cW = \mathbb{R}^{H \cdot d}$ the input noise for an $H$-step action chunk in a $d$-dimensional action space.

\subsection{Reward in Imagination}
\label{sec:app_reward}
Training inside the world model requires a reward that can be evaluated on imagined latent states, without
querying the environment. We therefore reuse the latent success classifier of DiWA \citep{chandra2025diwa},
a small network on top of the frozen world model encoder that predicts whether a latent state satisfies the
task. One classifier is trained per task on the demonstrations of that task, with states after task
completion as positives and a sample of the remaining states as negatives, and all classifiers stay frozen
during adaptation.

At every step of an imagined rollout on task $\ell_k$, the reward is the predicted class of that task's
classifier,\looseness=-1
\begin{equation}
\label{eq:app_reward}
r_t = \mathbb{1}\big[\,C_{\psi_k}(s_t) = 1\,\big] \in \{0, 1\},
\end{equation}
so the return in Eq.~\ref{eq:app_gae} counts the imagined steps in which the task is predicted solved. A
rollout also ends early once the classifier of the task being rolled out exceeds a confidence of $0.90$,
which keeps rollouts short and prevents the module from collecting reward past completion. The same
mechanism terminates rehearsal rollouts, using the classifier of the retained task rather than the current
one. Rehearsal itself uses no reward, since Eq.~\ref{eq:app_kl} only compares the module to its reference
on the visited states. On LIBERO no such classifier is used. The task is learned in the environment, where the reward is the
benchmark's own success signal, consumed by Eq.~\ref{eq:app_sac} in place of Eq.~\ref{eq:app_gae}. We do
not train a reward model in A2World's latent space, since the model is not accurate enough to optimize a
policy inside it (\S\ref{sec:sim_memory}), so rehearsal rollouts there run to a fixed length rather than
stopping at a predicted success.\looseness=-1

\subsection{Policy Optimization with PPO}
\label{sec:app_ppo_sub}
When the world model is faithful enough to act as a simulator (\S\ref{sec:two_roles}), the current task is
learned inside it. Imagined rollouts cost only forward passes through the frozen policy and world model and
run in parallel, so on-policy data is cheap and we optimize the module with PPO \citep{schulman2017ppo}.
This is the regime used on CALVIN and on the real robot. Each iteration collects a batch of imagined
rollouts with the current module, giving $\mathcal{D}_\text{cur}$, with rewards from Eq.~\ref{eq:app_reward}
normalized by a running estimate of their scale. On these rollouts we use generalized advantage estimation with the value function $V_\nu$,
\begin{equation}
\label{eq:app_gae}
\hat{A}_t = \sum_{l \geq 0} (\gamma \lambda_{\mathrm{GAE}})^l \, \delta_{t+l},
\qquad \delta_t = r_t + \gamma V_\nu(s_{t+1}) - V_\nu(s_t),
\qquad \hat{R}_t = \hat{A}_t + V_\nu(s_t),
\end{equation}
where $\hat{R}_t$ is the return target for the value loss. Both the current module and the one that
collected the rollouts are diagonal Gaussians over $\cW$, so the importance ratio
$\rho = \piw_k(\bw \mid s, \ell_k) / \piw_{\mathrm{old}}(\bw \mid s, \ell_k)$ is available in closed form and
\begin{equation}
\label{eq:app_ppo}
\mathcal{L}_\text{PPO} = - \mathbb{E}_{(s,\bw,\hat{A}) \sim \mathcal{D}_\text{cur}}
\Big[ \min\big( \rho \hat{A},\ \mathrm{clip}(\rho, 1-\epsilon, 1+\epsilon)\, \hat{A} \big) \Big].
\end{equation}
The value function is fit on the same batch,
\begin{equation}
\label{eq:app_value}
\mathcal{L}_V = \tfrac{1}{2}\, \mathbb{E}_{(s,\hat{R}) \sim \mathcal{D}_\text{cur}} \big[ (V_\nu(s) - \hat{R})^2 \big],
\end{equation}
and is not part of the module's objective in Eq.~\ref{eq:total}. Each batch is reused for several epochs,
stopping early once the KL between the updated module and the one that collected the rollouts exceeds a
threshold. Hyperparameters are listed in Appendix~\ref{sec:app_hparams}.

\subsection{Policy Optimization with SAC}
\label{sec:app_sac_sub}
When the world model is not accurate enough to optimize a policy inside it (\S\ref{sec:two_roles}), the
current task is learned in the environment instead. Every transition then costs a real episode, so the data
is scarce and has to be reused, and we train the module off-policy with SAC over the noise space, following
DSRL \citep{wagenmaker2025dsrl}. This is the regime used on LIBERO. Rehearsal is unaffected by the change, since
the KL term of Eq.~\ref{eq:app_kl} is still computed on rollouts imagined in the world model, so the robot
is used only for the current task.

In this regime the module is a tanh-squashed Gaussian over $\cW$, which keeps the sampled noise bounded
without the clipping used in imagination. A pair of critics $Q_{\nu_1}, Q_{\nu_2}$ on state and noise is fit
by temporal-difference learning on $\mathcal{D}_\text{cur}$, which here is the replay buffer of transitions
collected so far on the current task, and the actor maximizes the entropy-regularized value of its noise,
\begin{equation}
\label{eq:app_sac}
\begin{aligned}
\mathcal{L}_\text{SAC} &= \mathbb{E}_{s \sim \mathcal{D}_\text{cur},\ \bw \sim \piw_k(\cdot \mid s, \ell_k)}
\big[ \beta \log \piw_k(\bw \mid s, \ell_k) - \min_i Q_{\nu_i}(s, \bw) \big], \\[2pt]
\mathcal{L}_Q &= \sum_i \mathbb{E} \big[ (Q_{\nu_i}(s, \bw) - r - \gamma \bar{V}(s'))^2 \big], \\[2pt]
\bar{V}(s') &= \mathbb{E}_{\bw' \sim \piw_k} \big[ \min_i \bar{Q}_{\nu_i}(s', \bw') - \beta \log \piw_k(\bw' \mid s', \ell_k) \big],
\end{aligned}
\end{equation}
where the target critics $\bar{Q}_{\nu_i}$ are updated by Polyak averaging. The temperature $\beta$ is tuned
automatically against a target entropy of $-|\cW|/2$ rather than held fixed. Hyperparameters are listed in Appendix~\ref{sec:app_hparams}.

\subsection{Behavior-Cloning Loss}
\label{sec:app_bc_sub}
Optimizing against a learned model invites the module to find noise that the model rewards but the robot
does not. Following DiWA \citep{chandra2025diwa}, we therefore anchor the steered actions to the ones the
frozen policy produces on its own. The anchor costs nothing to obtain, since drawing noise from the prior
recovers the unsteered policy, so no teacher network has to be stored alongside the module. With
$\mu_k(s,\ell_k)$ the mean of the steering module,
\begin{equation}
\label{eq:app_bc}
\mathcal{L}_\text{BC} = \mathbb{E}_{s \sim \mathcal{D}_\text{cur},\ \bw_0 \sim \mathcal{N}(0,I)}
\Big[ \big\lVert \pidpw\big(s, \ell_k, \mu_k(s,\ell_k)\big) - \pidpw(s, \ell_k, \bw_0) \big\rVert_2^2 \Big],
\end{equation}
computed over the executed steps of the chunk and averaged over the current-task states. The term is only needed where the reward comes from the world model. When the task is learned in the
environment the reward is real, so there is nothing to exploit and $\alpha = 0$ in Eq.~\ref{eq:total}. Its
value in imagination is listed in Table~\ref{tab:app_hparams}, and \S\ref{sec:app_alpha} reports how
performance depends on it.

\subsection{Rehearsal Loss}
\label{sec:app_kl_sub}
Rehearsal is what prevents earlier skills from drifting while a new task is learned. It is applied the same
way in both regimes, since the retained tasks are always rolled out inside the frozen world model and the
robot is never revisited for a task that is not being learned. 

For each retained task $\ell_j \in \mathcal{R}_k$, the states $\smash{\mathcal{D}^{j}_{\mathrm{reh}}}$ come
from rollouts seeded at a stored initial observation of $\ell_j$ and unrolled with the reference
$\smash{\piw_{\mathrm{ref}(j)}}$ choosing the noise, so they follow the behavior that is to be preserved.
The reference is the previous module $\smash{\piw_{k-1}}$ for tasks that were already steered and the prior
$\mathcal{N}(0,I)$ for the rest. On CALVIN these rollouts are regenerated at every iteration rather than
stored, and since the world model is frozen their distribution does not drift as the sequence grows. The penalty is Eq.~\ref{eq:kl} in the main text, averaged over the retained tasks with a single coefficient
$\lambda$. Both distributions are diagonal Gaussians, so the divergence of the current module from the
reference is available in closed form,
\begin{equation}
\label{eq:app_kl}
\mathrm{KL}\big(\mathcal{N}(\mu,\sigma^2)\,\|\,\mathcal{N}(\mu',\sigma'^2)\big) = \tfrac{1}{2} \sum_i \Big[ \frac{\sigma_i^2 + (\mu_i - \mu'_i)^2}{\sigma'^2_i} - 1 + \log \sigma'^2_i - \log \sigma_i^2 \Big],
\end{equation}
with $\mu' = 0$ and $\sigma' = 1$ when the reference is the prior. The value of $\lambda$ is listed in
Table~\ref{tab:app_hparams}, and \S\ref{sec:app_lambda} reports how performance depends on it.

\section{Algorithm}
\label{sec:app_algorithm}
Algorithm~\ref{alg:d2r} summarizes one stage of D2R. Each iteration collects experience on the current
task, regenerates rehearsal rollouts for the retained tasks inside the world model, and updates the module
on the combined objective of Eq.~\ref{eq:total}, with the choice of MDP and reinforcement learning algorithm
following \S\ref{sec:two_roles}.

\begin{algorithm}[p]
\caption{D2R: Continual Steering of a Frozen Generative Policy with World-Model Rehearsal}
\label{alg:d2r}
\begin{algorithmic}[1]
\State \textbf{Input:} frozen policy $\pidpw$, frozen world model $P_\phi$, task rewards $R_j$, task sequence $\ell_1, \dots, \ell_N$, initial observations $\mathcal{O}_j$ from the demonstrations of each task $j$.
\State Set the prior $\piw_{\mathrm{prior}} \gets \mathcal{N}(0, I)$, the unsteered policy's noise distribution.
\For{task $k = 1, \dots, N$ in sequence}
    \State Freeze the previous module $\piw_{k-1}$.
    \State Initialize a new steering module $\piw_k$ from scratch.
    \State Initialize the critic of the chosen RL algorithm ($V_\nu$ for PPO, $Q_\nu$ for SAC).
    \State Choose the MDP for task $k$, $\Mwm$ if the world model covers the task and $\Menv$ otherwise.
    \State Define the retention set $\mathcal{R}_k$, the tasks to be rehearsed while learning task $k$.
    \State For each $j \in \mathcal{R}_k$, set its reference $\mathrm{ref}(j)$ to $\piw_{k-1}$ if task $j$ was steered before, else to $\piw_{\mathrm{prior}}$.
    \For{iteration $= 1, 2, \dots$}
        \State Initialize buffers $\mathcal{D}_\text{cur}$ and $\mathcal{D}_\text{reh}$ ($\mathcal{D}_\text{cur}$ persists for off-policy algorithms).
        \Statex \vspace{-0.6em}\hrulefill
        \State \textbf{1. Adaptation}
        \For{episode $= 1, \dots, N_\text{cur}$ in parallel}
            \State Sample initial observation $o_0 \sim \mathcal{O}_k$ and encode it to $s_0$.
            \For{step $t = 0, \dots, T-1$}
                \State Sample noise $\bw_t \sim \piw_k(\cdot \mid s_t, \ell_k)$ (Eq.~\ref{eq:steer}).
                \State Decode action chunk $\ba_t \gets \pidpw(s_t, \ell_k, \bw_t)$.
                \If{task $k$ is learned in $\Mwm$}
                    \State Step the world model: $s_{t+1} \sim P_\phi(\cdot \mid s_t, \ba_t)$.
                \Else
                    \State Step the environment with $\ba_t$, observe $o_{t+1}$, and encode it to $s_{t+1}$.
                \EndIf
                \State Observe reward $r_t = R_k(s_{t+1})$.
                \State Add $(s_t, \bw_t, r_t)$ to $\mathcal{D}_\text{cur}$.
            \EndFor
        \EndFor
        \Statex \vspace{-0.6em}\hrulefill
        \State \textbf{2. Rehearsal}
        \For{rehearsed task $j \in \mathcal{R}_k$, episode $= 1, \dots, N_\text{reh}$ in parallel}
            \State Sample initial observation $o_0 \sim \mathcal{O}_j$ and encode it to $s_0$.
            \For{step $t = 0, \dots, T-1$}
                \State Set noise to the reference mean: $\bw_t \gets \mathrm{mean}\big[\piw_{\mathrm{ref}(j)}(\cdot \mid s_t, \ell_j)\big]$.
                \State Decode action chunk $\ba_t \gets \pidpw(s_t, \ell_j, \bw_t)$.
                \State Step the world model: $s_{t+1} \sim P_\phi(\cdot \mid s_t, \ba_t)$.
                \State Add $(s_t, j)$ to $\mathcal{D}_\text{reh}$.
                \State Terminate on success if a detector exists, otherwise at step $T$.
            \EndFor
        \EndFor
        \Statex \vspace{-0.6em}\hrulefill
        \State \textbf{3. Update}
        \State For on-policy algorithms, compute GAE advantages and returns on $\mathcal{D}_\text{cur}$ (Eq.~\ref{eq:app_gae}).
        \For{update $= 1, \dots, U$}
            \For{minibatch $= 1, \dots, B$}
                \State Sample a current-task minibatch from $\mathcal{D}_\text{cur}$ and rehearsal states $(s, j)$ from $\mathcal{D}_\text{reh}$.
                \State Compute the RL loss $\mathcal{L}_\text{RL}$, e.g.\ PPO (Eq.~\ref{eq:app_ppo}) or SAC (Eq.~\ref{eq:app_sac}).
                \State Compute the rehearsal loss $\mathcal{L}_\text{KL}$ on $\mathcal{D}_\text{reh}$ against $\piw_{\mathrm{ref}(j)}$ (Eq.~\ref{eq:kl}).
                \State Compute the BC loss $\mathcal{L}_\text{BC}$ if task $k$ is learned in $\Mwm$, else $0$ (Eq.~\ref{eq:app_bc}).
                \State Update $\piw_k$ on $\mathcal{L}_\text{RL} + \alpha \mathcal{L}_\text{BC} + \lambda \mathcal{L}_\text{KL}$ (Eq.~\ref{eq:total}).
                \State Update the critic on its loss (Eq.~\ref{eq:app_value} or Eq.~\ref{eq:app_sac}).
            \EndFor
        \EndFor
    \EndFor
\EndFor
\State \Return $\piw_N$, deployed on every task. Disabling it recovers $\pidp$.
\end{algorithmic}
\end{algorithm}

\section{Implementation Details}
\label{sec:app_impl}
This section describes the components D2R builds on and the one it trains. The world models, base policies,
and reward classifiers stay frozen throughout, and only the steering module and its critic are learned.

\subsection{World Models}
On CALVIN and on the real robot we use the recurrent state-space model of DiWA \citep{chandra2025diwa} and
LUMOS \citep{nematollahi2025lumos} without modification, trained once on task-agnostic play data and frozen
thereafter. It encodes a static and a gripper camera view into a latent state and predicts the next latent
from an action chunk, and we refer to those works for the architecture and the training objective. The
CALVIN model is trained on play data from environment~\textit{D}, and the real-robot model on six hours of
teleoperated play collected on the same Franka, with the same architecture and settings. On LIBERO we use A2World \citep{huang2026learning} off the shelf, pretrained on 2.1M real-robot
trajectories and fine-tuned on the four suites, so that neither it nor $\pi_0$ has seen the LIBERO-90 target
tasks. It is likewise frozen.

The two regimes use these models differently. On CALVIN and the real robot the model supplies both the
rollouts the module learns in and the latent states it conditions on. On LIBERO it supplies only rehearsal
rollouts, and the module never reads an A2World latent, conditioning instead on the raw observation
(Table~\ref{tab:app_hparams}).

\subsection{Base Policies}
\label{sec:app_base_policies}
On CALVIN and on the real robot the base policy is a single language-conditioned diffusion policy trained by
behavior cloning, following DiWA \citep{chandra2025diwa}. It differs from DiWA in covering all tasks at
once rather than one policy per skill, and it conditions on the same world-model latents the module later
sees in imagination, with the instruction embedding fused into those features by FiLM conditioning
\citep{perez2018film}. The CALVIN policy is trained on 150 demonstrations for each of 20 CALVIN tasks,
eight of which form the adaptation sequence, and the real-robot policy on 100 demonstrations for each of the
seven tabletop tasks. On LIBERO the base policy is the publicly released $\pi_0$ checkpoint \citep{black2024pi_0} fine-tuned on
the Spatial, Object, Goal, and Long suites, which we use as is.

What the steering module controls in all three cases is the initial noise the sampler starts from, so the
sampler itself is left untouched, DDIM on CALVIN and the real robot and the flow ODE of $\pi_0$ on LIBERO,
with step counts and embedding dimensions in Table~\ref{tab:app_hparams}. All base policies remain frozen
throughout.

\subsection{Rewards}
The latent success classifier is two small multilayer perceptrons on the frozen $2048$-dimensional
world-model latent, an encoder to a $256$-dimensional embedding and a two-way classification head, both
with a hidden width of $1024$. It is trained with the contrastive and cross-entropy objective of DiWA
\citep{chandra2025diwa}, to which we refer for the objective itself and for the comparison against a
vision-based classifier. We reuse the classifiers released with DiWA rather than retraining
them. On the real robot one classifier is trained per task on the same demonstrations as the base policy.
No classifier is trained for A2World, for the reason given in \S\ref{sec:app_reward}.

\subsection{Steering Module and Critics}
\label{sec:app_module}
The steering module is a three-layer perceptron of width $512$ with layer normalization that outputs a
diagonal Gaussian over $\cW$, with a state-dependent standard deviation bounded to the range in
Table~\ref{tab:app_hparams}. What it conditions on differs by benchmark. On CALVIN and the real robot it
reads the world-model latent, while on LIBERO it encodes the raw observation with its own encoder, and in
both cases the instruction embedding is fused into those features
by FiLM conditioning, as in the base policy. Samples are clipped in imagination and
tanh-squashed in the environment (\S\ref{sec:app_sac_sub}).

The critic follows the algorithm of the regime, a single value network in imagination and a pair of
$Q$~networks in the environment, with the widths listed in Table~\ref{tab:app_hparams}. Both the module and
its critic are initialized from scratch at every stage, and the critic is discarded once the stage ends, so
the module is the only component carried forward.
\subsection{Hyperparameters}
\label{sec:app_hparams}
Table~\ref{tab:app_hparams} lists the hyperparameters of D2R in both regimes. The frozen components keep
the settings of the works they come from. On CALVIN the module emits one value per action dimension and
executed step, so $|\cW| = H \cdot d$. On LIBERO it emits a single $d$-dimensional vector that the sampler
applies across the whole chunk.

\begin{table}[H]
\footnotesize
\centering
\caption{Hyperparameters of D2R. CALVIN adapts in imagination with PPO, LIBERO adapts in the environment with SAC. Rehearsal runs in the world model in both cases. The real robot follows the CALVIN column.}
\label{tab:app_hparams}
\setlength{\tabcolsep}{4pt}
\begin{tabular}{llll}
\toprule
\textbf{Parameter} & \textbf{Symbol} & \textbf{CALVIN} & \textbf{LIBERO} \\
\midrule
\multicolumn{4}{l}{\textit{Frozen policy and noise space}} \\
Action chunk length & $H$ & 4 & 50 \\
Executed actions & --- & 4 & 20 \\
Action dimension & $d$ & 7 & 32 \\
Noise dimension & $|\cW|$ & 28 & 32 \\
Sampler & --- & DDIM & flow ODE \\
Sampling steps & --- & 10 of 20 & 10 of 10 \\
\midrule
\multicolumn{4}{l}{\textit{Steering module}} \\
State input & $s$ & world-model feature (2048) & input image (64x64) \\
Instruction input & $\ell$ & CLIP embedding (512) & PaliGemma embedding (2048) \\
Hidden layers & --- & [512, 512, 512], LayerNorm & [512, 512, 512] \\
Std bounds & $[\sigma_{\min},\sigma_{\max}]$ & [0.1, 1.5] & [0, 1] \\
Sample clipping & $c$ & [-3, 3] & [-1, 1] \\
Initialization per task & --- & from scratch & from scratch \\
\midrule
\multicolumn{4}{l}{\textit{Adaptation}} \\
Algorithm & --- & PPO & SAC \\
Iterations / env.\ steps & --- & 800 iterations & 250k grad.\ steps \\
Parallel rollouts & --- & 50 & 1 \\
Rollout length (chunks) & $T$ & 36 & --- \\
Discount & $\gamma$ & 0.999 & 0.999 \\
GAE parameter & $\lambda_{\mathrm{GAE}}$ & 0.95 & --- \\
PPO clip & $\epsilon$ & 0.01 & --- \\
Update epochs per iteration & --- & 10 & --- \\
Batch size & --- & 7500 & 256 \\
Actor / critic learning rate & --- & $10^{-4}$ / $10^{-3}$ & $10^{-4}$ / $3\times   10^{-4}$ \\
Critic hidden layers & --- & [256, 256, 256] & [512, 512, 512] \\
Entropy temperature & $\beta$ & --- & learned, $\beta_0 = 1$ \\
Update-to-data ratio & --- & --- & 20 \\
\midrule
\multicolumn{4}{l}{\textit{Rehearsal}} \\
Seed frames per task & --- & 150 demo initial frames & 50 demo initial frames \\
Parallel rollouts per task & --- & 10 & 1 \\
Rollout length (chunks) & $T$ & 36 & 200 \\
Early termination & --- & classifier $\geq 0.9$ & none \\
\midrule
\multicolumn{4}{l}{\textit{Loss weights (Eq.~\ref{eq:total})}} \\
Behavior cloning & $\alpha$ & $5\times10^{-2}$ & 0 \\
Retention & $\lambda$ & $5\times10^{-3}$ & $5\times10^{-3}$ \\
\bottomrule
\end{tabular}
\end{table}

\subsection{Compute and Wall-Clock}
\label{sec:app_compute}
All adaptation runs use a single A100 GPU, except for the rehearsal rollouts on LIBERO, which
are generated across 5 GPUs before each stage. One training stage takes about an
hour on CALVIN, eight hours on LIBERO, and 30 minutes on the real robot.

Rehearsal is the only cost D2R adds over steering without it. On CALVIN, where the world model and the base
policy are both small, the retained-task rollouts are regenerated at every iteration, which lengthens a stage
by 25\% relative to DSRL under the same budget. On LIBERO they are instead generated once at
the start of a stage, since every imagined step queries both A2World and $\pi_0$ and placing that inside the
update loop would dominate the wall clock. Generating them takes 30 minutes for
50 episodes per retained task, which is amortized over the gradient steps that follow. Since the
world model, the base policy, and the reward classifiers are frozen and reused across the sequence, the figures above exclude their training.

\section{Experimental Setup Details}
\label{sec:app_setup}
\subsection{Environments and Task Sequences}

\begin{figure}[h]
    \centering
    \input{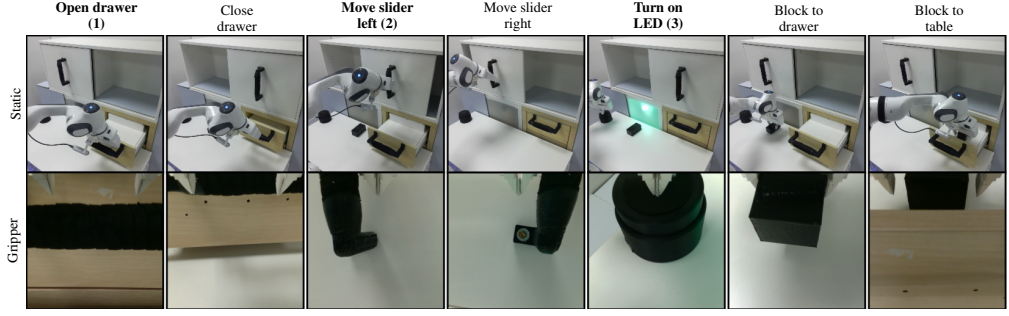}
    \caption{\textbf{The seven real-robot tasks.} Each column is one task, seen by the static camera (top)
    and the gripper camera (bottom). The three tasks adapted in sequence are set in bold with their stage
    number.}
    \label{fig:app_real_skills}
\end{figure}

Table~\ref{tab:task_instructions} lists the tasks and their instructions for the CALVIN and LIBERO benchmarks. On the Franka robot we use seven tabletop tasks, open drawer, close drawer, move slider left, move slider right, turn on LED, pick the block from the table and place it in the drawer, and pick the block from the drawer and place it on the table, all shown in Figure~\ref{fig:app_real_skills}. Five of them also appear in CALVIN, while the two block tasks do not. The three adapted tasks are open drawer, move slider left, and turn on LED, learned in this order.

\begin{table*}[t]
    \centering
    \caption{Task instructions used in CALVIN and LIBERO. The real tasks are
    shown in Figure~\ref{fig:app_real_skills}.}
    \label{tab:task_instructions}

    \renewcommand{\arraystretch}{1.15}
    \setlength{\tabcolsep}{4pt}
    \footnotesize

\begin{tabularx}{\textwidth}{
    >{\centering\arraybackslash}p{0.045\textwidth}
    >{\raggedright\arraybackslash}X
    |
    >{\centering\arraybackslash}p{0.045\textwidth}
    >{\raggedright\arraybackslash}X
}
\toprule

% =========================================================
% CALVIN and LIBERO-90
% =========================================================
\multicolumn{2}{c|}{\textbf{CALVIN}}
&
\multicolumn{2}{c}{\textbf{LIBERO-90}}
\\
\cmidrule(lr){1-2}\cmidrule(lr){3-4}

\textbf{ID} & \textbf{Task Instruction}
&
\textbf{ID} & \textbf{Task Instruction}
\\
\midrule

1 & Close the drawer.
& 0 & Close the top drawer of the cabinet. \\

2 & Move the slider to the left.
& 7 & Open the top drawer of the cabinet. \\

3 & Move the slider to the right.
& 9 & Put the black bowl on the plate. \\

4 & Open the drawer.
& 10 & Put the black bowl on top of the cabinet. \\

5 & Turn off the LED.
& 56 & Pick up the butter and put it in the tray. \\

6 & Turn off the lightbulb.
& 57 & Pick up the cream cheese and put it in the tray. \\

7 & Turn on the LED.
& 20 & Turn on the stove. \\

8 & Turn on the lightbulb.
& 35 & Open the microwave. \\

\midrule

% =========================================================
% LIBERO-Spatial and LIBERO-Goal
% =========================================================
\multicolumn{2}{c|}{\textbf{LIBERO-Spatial}}
&
\multicolumn{2}{c}{\textbf{LIBERO-Goal}}
\\
\cmidrule(lr){1-2}\cmidrule(lr){3-4}

\textbf{ID} & \textbf{Task Instruction}
&
\textbf{ID} & \textbf{Task Instruction}
\\
\midrule

0 & Pick up the black bowl between the plate and the ramekin and place it on the plate.
& 0 & Open the middle drawer of the cabinet. \\

1 & Pick up the black bowl next to the ramekin and place it on the plate.
& 1 & Put the bowl on the stove. \\

2 & Pick up the black bowl from the table center and place it on the plate.
& 2 & Put the wine bottle on top of the cabinet. \\

3 & Pick up the black bowl on the cookie box and place it on the plate.
& 3 & Open the top drawer and put the bowl inside. \\

4 & Pick up the black bowl in the top drawer of the wooden cabinet and place it on the plate.
& 4 & Put the bowl on top of the cabinet. \\

5 & Pick up the black bowl on the ramekin and place it on the plate.
& 5 & Push the plate to the front of the stove. \\

6 & Pick up the black bowl next to the cookie box and place it on the plate.
& 6 & Put the cream cheese in the bowl. \\

7 & Pick up the black bowl on the stove and place it on the plate.
& 7 & Turn on the stove. \\

8 & Pick up the black bowl next to the plate and place it on the plate.
& 8 & Put the bowl on the plate. \\

9 & Pick up the black bowl on the wooden cabinet and place it on the plate.
& 9 & Put the wine bottle on the rack. \\

\midrule

% =========================================================
% LIBERO-Object and LIBERO-10
% =========================================================
\multicolumn{2}{c|}{\textbf{LIBERO-Object}}
&
\multicolumn{2}{c}{\textbf{LIBERO-Long}}
\\
\cmidrule(lr){1-2}\cmidrule(lr){3-4}

\textbf{ID} & \textbf{Task Instruction}
&
\textbf{ID} & \textbf{Task Instruction}
\\
\midrule

0 & Pick up the alphabet soup and place it in the basket.
& 0 & Put both the alphabet soup and the tomato sauce in the basket. \\

1 & Pick up the cream cheese and place it in the basket.
& 1 & Put both the cream cheese box and the butter in the basket. \\

2 & Pick up the salad dressing and place it in the basket.
& 2 & Turn on the stove and put the moka pot on it. \\

3 & Pick up the BBQ sauce and place it in the basket.
& 3 & Put the black bowl in the bottom drawer of the cabinet and close it. \\

4 & Pick up the ketchup and place it in the basket.
& 4 & Put the white mug on the left plate and put the yellow and white mug on the right plate. \\

5 & Pick up the tomato sauce and place it in the basket.
& 5 & Pick up the book and place it in the back compartment of the caddy. \\

6 & Pick up the butter and place it in the basket.
& 6 & Put the white mug on the plate and put the chocolate pudding to the right of the plate. \\

7 & Pick up the milk and place it in the basket.
& 7 & Put both the alphabet soup and the cream cheese box in the basket. \\

8 & Pick up the chocolate pudding and place it in the basket.
& 8 & Put both moka pots on the stove. \\

9 & Pick up the orange juice and place it in the basket.
& 9 & Put the yellow and white mug in the microwave and close it. \\

\bottomrule
\end{tabularx}

\end{table*}
\subsection{Real-World Setup}
\label{sec:app_real_setup}
\begin{figure}[h]
    \centering
    \includegraphics[width=0.5\textwidth]{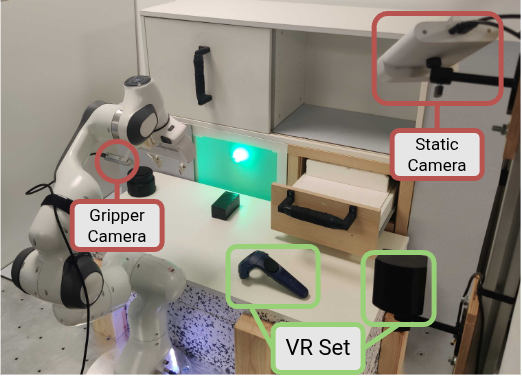}
    \caption{\textbf{Real-world setup.} A Franka Emika Panda in a tabletop environment with a cabinet, a
    drawer, a sliding door, and an LED button. The policy observes the scene through a static camera and a
    gripper-mounted camera. Play data and demonstrations are teleoperated with a VR controller.}
    \label{fig:app_robot_env}
\end{figure}

Experiments on hardware use a 7-DoF Franka Emika Panda in the tabletop environment of
Figure~\ref{fig:app_robot_env}, which mirrors the CALVIN scene with a cabinet, a drawer, a sliding door, an
LED button, and a movable block. The policy observes the scene through a static camera placed over the
shoulder of the robot and a camera mounted on the gripper, both at $64\times64$ resolution, and acts through
the same 7-dimensional end-effector command as in simulation (\S\ref{sec:app_action_space}). All play data
and demonstrations are teleoperated through a VR controller.

Three components are learned once on this platform and then frozen. The world model is an RSSM trained on six
hours of teleoperated play data. The base policy is a single language-conditioned diffusion policy trained on
100 demonstrations for each of the seven tasks, and each task additionally has a success classifier trained
on the same demonstrations. D2R and DSRL are then trained entirely in imagination, so the robot is used only
for evaluation. After each stage we run 20 episodes per adapted task of at most 72 steps,
starting from 20 fixed initial configurations that are reset by hand between episodes.

\subsection{Action Space}
\label{sec:app_action_space}
All experiments, in simulation and on the real robot, use the same 7-dimensional relative action space,\looseness=-1
\begin{equation}
\label{eq:app_action}
\mathbf{a} = [\delta x, \delta y, \delta z, \delta \phi, \delta \theta, \delta \psi, a_{\mathrm{gripper}}].
\end{equation}
The first three entries are changes in end-effector position and the next three are changes in its
orientation, expressed as Euler angles, each taking continuous values in $[-1, 1]$. The last entry commands
the gripper. On LIBERO, $\pi_0$ emits a 32-dimensional action vector, of which only the first seven
dimensions are sent to the robot and the rest are discarded. The steering module still acts on the full
32-dimensional noise vector of the sampler, which Table~\ref{tab:app_hparams} reports as the noise
dimension.\looseness=-1

\subsection{Rehearsal Seed Frames and Storage Cost}
\label{sec:app_seeds}
Rehearsal needs a starting point for every retained task, and these starting points are the only thing D2R
keeps from a task once its stage is over. For each task we store one frame per demonstration, taken from the
first step of that demonstration, extracted once before the sequence begins and never updated afterwards. On
CALVIN a frame is kept as its 2048-dimensional world-model feature, and on LIBERO as the initial
observation of the benchmark episode. The demonstrations are drawn at random under a fixed seed.
Rollouts are then generated from these frames inside the frozen world model as described in
\S\ref{sec:app_kl_sub}, so no trajectory of a retained task is ever stored or revisited.

The storage column of Table~\ref{tab:sequence} compares this against a replay buffer that keeps the
retained-task trajectories themselves, which is what the replay KL baseline requires. Since we keep one frame
per demonstration, the two hold the same episodes and differ only in how much of each episode they keep, so
\begin{equation}
\label{eq:app_storage}
\text{replay frames per task} \;=\; (\text{seed frames per task}) \times \bar{T},
\end{equation}
where $\bar{T}$ is the mean length of those demonstrations, 60.2 frames on CALVIN and 148.7 frames on the
LIBERO-90 tasks that make up the sequence. After the full eight-task sequence the two hold

\begin{center}
\footnotesize
\setlength{\tabcolsep}{6pt}
\begin{tabular}{lrr}
\toprule
\textbf{Kept after the sequence} & \textbf{CALVIN} & \textbf{LIBERO} \\
\midrule
Seed frames per task & 150 & 50 \\
Seed frames, eight tasks & 1{,}200 & 400 \\
\midrule
Mean episode length $\bar{T}$ & 60.2 & 148.7 \\
Replay frames per task & 9{,}030 & 7{,}435 \\
Replay frames, eight tasks & 72{,}240 & 59{,}480 \\
\bottomrule
\end{tabular}
\end{center}

\noindent and Table~\ref{tab:sequence} reports the LIBERO figures, about 400 frames against about
59{,}000. The gap is exactly the trajectory dimension, since the ratio on each benchmark is its mean episode
length, 60 on CALVIN and 149 on LIBERO. What D2R keeps grows only with the number of retained tasks, whereas
a replay buffer grows with the length of every stored episode as well, so the ratio widens as tasks get
longer.
\subsection{Baselines}
\label{sec:app_baselines}
The baselines are introduced in \S\ref{sec:sim_setup}. This section gives only what is held fixed across
them and the settings in which they differ. All of them start from the same frozen base policy and use the
same world model, reward classifiers, seed frames, task order, and evaluation protocol, and those that train do
so under the same per-stage budget. Any hyperparameter not named below is the one D2R uses in
Table~\ref{tab:app_hparams}.

\textbf{DiWA.}~The published method \citep{chandra2025diwa}, which adapts by fine-tuning all policy weights in imagination with PPO rather than by steering the noise. Since every method starts from the same base policy, it fine-tunes the multi-task policy of \S\ref{sec:app_base_policies} rather than the per-skill policies of the released setup.

\textbf{DiWA + rehearsal.}~DiWA with the world-model rehearsal of D2R added, run in single-task adaptation on CALVIN only. The policy weights $\theta$ are fine-tuned on the adapted task while the other seven are rehearsed on the same imagined rollouts D2R uses. Since the fine-tuned policy has no tractable action distribution, the KL penalty of Eq.~\ref{eq:kl} is replaced by an action-level anchor to the frozen base policy on the rehearsal states,
\begin{equation}
\label{eq:app_diwa_reh}
\mathcal{L}_\text{reh}^{\text{DiWA}} = \sum_{j \neq k}\ \mathbb{E}_{s \sim \mathcal{D}^{j}_\text{reh},\ \bw \sim \mathcal{N}(0,I)}
\Big[ \big\lVert \pi_\theta(s, \ell_j, \bw) - \pidpw(s, \ell_j, \bw) \big\rVert_2^2 \Big],
\end{equation}
with the same noise fed to both policies, weighted by the same coefficient $\lambda$ as D2R. It is not run in the sequence experiments, where LoRA fine-tuning is the weight-space method, because fine-tuning all weights at every stage is the setting DiWA itself does not target.

\textbf{LoRA fine-tuning.}~On CALVIN, DiWA's PPO fine-tuning in imagination with the policy weights frozen and LoRA adapters \citep{hu2022lora} of rank 64 with scaling factor 64, no dropout, injected into every linear layer of the policy's denoising network. Training follows the schedule of D2R apart from the actor learning rate, $10^{-5}$. On LIBERO, adapters of rank 8 with scaling factor 8 are applied only to the action expert of $\pi_0$, across its 18 transformer layers, with the vision-language backbone frozen, and trained off policy in the environment. We tried both SAC and an IQL-style objective and report the latter, which performed better, since $\pi_0$ exposes no exact action log-probability and the entropy term of SAC has no closed form. It uses twin Q networks, a value network with expectile $0.7$, and an advantage-weighted flow-matching actor update with behavior-cloning coefficient $0.1$ and an anchor to the initial adapters.

\textbf{DP-TTC.}~Test-time selection on CALVIN, which trains nothing and instead queries the world model and the success classifier at every decision. It draws 25 candidate chunks at temperature 1, imagines each one chunk ahead, and executes one drawn uniformly from the 10 best by the classifier's predicted change in success, without refining the candidates.

\textbf{DSRL.}~The steering approach of \citet{wagenmaker2025dsrl} without rehearsal, so the module of \S\ref{sec:app_module} is trained on the current task alone and Eq.~\ref{eq:total} is used with $\lambda = 0$. It follows the regime of D2R on each benchmark, PPO in the world model on CALVIN and SAC in the environment on LIBERO, which departs from the released setup where steering is always learned from environment interaction.

The three remaining input-space baselines add the rehearsal penalty of Eq.~\ref{eq:kl} to DSRL with the same coefficient $\lambda$ as D2R and differ only in the states it is applied to, so each is described by that set alone.\looseness=-1

\textbf{DSRL + instruction KL.}~The penalty is applied to observations of the current task paired with instructions of the retained tasks, so no state of a retained task is ever stored or generated. On LIBERO the mismatched instruction is drawn from the four suites $\pi_0$ was fine-tuned on, one per observation.\looseness=-1

\textbf{DSRL + first-obs KL.}~The penalty is applied to the stored seed frames of \S\ref{sec:app_seeds} directly, without rolling them forward in the world model, so it sees only states in which a retained task has not yet begun.\looseness=-1

\textbf{DSRL + replay KL.}~The penalty is applied to rollouts of the retained tasks collected once at the start of each stage with the previous module and stored for the rest of it, real trajectories on LIBERO and imagined ones on CALVIN. It differs from D2R on CALVIN only in that the rollouts are stale rather than regenerated at every iteration.\looseness=-1

The last two rows of Table~\ref{tab:sequence} are reference settings rather than baselines. Both use the steering module and the objective of D2R, and both remove the sequence in a different way, so together they bracket what sequential steering can reach.\looseness=-1

\textbf{Per-task modules.}~One module is trained for each task and selected at evaluation by the task identity. Each module therefore sees only its own task, nothing has to be retained, and Eq.~\ref{eq:total} is used with $\lambda = 0$. This is the ceiling for steering at the same module capacity, paid for by storing one module per task and by requiring the task identity at test time, neither of which D2R needs.

\textbf{Multi-task module.}~A single module is trained on all eight tasks at once rather than in sequence, with a larger budget than any single stage of D2R, namely longer training, more collected experience, and more parallel environments, at the same module capacity. Every task is present throughout, so again nothing has to be retained and $\lambda = 0$. Retention and forgetting are undefined here because no task has a stage at which it was learned, which is why those cells are marked in Table~\ref{tab:sequence}.

\subsection{Metrics}
\label{sec:app_metrics}
Tasks arrive in the order $\ell_1,\dots,\ell_N$. After the stage on task $\ell_k$ we evaluate the resulting system on every task and record its success rate $S_{k,j}$ on task $\ell_j$, in percent. This gives a success matrix $S$ whose rows are the stages $0,\dots,N$ and whose columns are the evaluated tasks, where row $0$ holds the success $S_{0,j}$ of the frozen base policy. The three metrics of Table~\ref{tab:sequence} are functions of this matrix.

\textbf{Final average success.}~The success of the system obtained after the last stage, averaged over the sequence, is the mean of the last row,
\begin{equation}
\text{Final avg.} \;=\; \frac{1}{N}\sum_{j=1}^{N} S_{N,j},
\label{eq:final_avg}
\end{equation}
reported as Success in Table~\ref{tab:sequence}. It coincides with the AVG metric of \citet{hu2026continualvlarl}. For the base policy the last row equals $S_{0,\cdot}$, so its entry is the base average.

\textbf{Forgetting.}~Forgetting measures how much success on a task drops after later tasks are learned. Since each task is learned once, we measure the drop relative to the success right after the task's own stage and average over the tasks learned before the last stage,
\begin{equation}
\text{Forgetting} \;=\; \frac{1}{N-1}\sum_{j=1}^{N-1}\left(S_{j,j}-S_{N,j}\right),
\label{eq:forgetting}
\end{equation}
which is the negative backward transfer (NBT) of \citet{hu2026continualvlarl} and the backward transfer of \citet{lopezpaz2017gem} with opposite sign. The NBT of \citet{liu2023libero} averages the same drop over every later stage rather than taking it at the end. More generally, $S_{j,j}-S_{k,j}$ is the forgetting score on task $\ell_j$ after stage $k$. Positive values are losses, and zero means no forgetting on average. The last task is excluded because nothing is learned after it. Task-specific modules never change after their own stage, so their forgetting is zero by construction.

\textbf{Retention.}~Retention measures the change in success relative to the frozen base policy, so it is defined for tasks that were never a stage of the sequence as well. After stage $k$ the retention on task $\ell_j$ is $S_{k,j}-S_{0,j}$. Table~\ref{tab:sequence} reports its mean after the last stage over the retention set $\mathcal{R}$,
\begin{equation}
\text{Retention} \;=\; \frac{1}{|\mathcal{R}|}\sum_{j\in\mathcal{R}}\left(S_{N,j}-S_{0,j}\right),
\label{eq:retention}
\end{equation}
where $\mathcal{R}$ is the eight tasks of the sequence on CALVIN and the four suites on LIBERO that $\pi_0$ was fine-tuned on and that never appear in the sequence. On CALVIN this equals Success minus the base average. On LIBERO it is the change in the held-out success ZS of \citet{hu2026continualvlarl} relative to $\pi_0$. In single-task adaptation (Table~\ref{tab:calvin_results_v2}) there is no sequence. Writing $S^{(i)}_{j}$ for the success on task $\ell_j$ after adapting to task $\ell_i$ alone, the retention reported for task $\ell_i$ is the mean change over all tasks,
\begin{equation}
\text{Retention}^{(i)} \;=\; \frac{1}{N}\sum_{j=1}^{N}\left(S^{(i)}_{j}-S_{0,j}\right),
\label{eq:retention_single}
\end{equation}
including the adapted task itself, so that gains on $\ell_i$ and losses on the other tasks enter with their sign. Positive values mean the adapted system ends above the base policy, negative values mean it ends below.
\section{Additional Results and Ablations}
\label{sec:app_ablations}

\subsection{Full Sequential Results}
\label{sec:app_sequence_full}
Table~\ref{tab:app_sequence_full} repeats Table~\ref{tab:sequence} with two further columns per benchmark. \emph{Single-task} is the average success when each task is adapted on its own, outside the sequence, and \emph{Current-task} is the mean of the diagonal $\frac{1}{N}\sum_{j} S_{j,j}$, the success on each task immediately after its own adaptation stage. The difference between Single-task and Current-task is the cost of learning a task inside the sequence rather than in isolation, and the difference between Current-task and Success is $(N-1)/N$ times the forgetting of Eq.~\ref{eq:forgetting}, which averages the same per-task drops but excludes the last task. A module trained on a single task has nothing to retain, so the Single-task column of DSRL is the per-task modules setting of Table~\ref{tab:sequence}.
\begin{center}\begin{minipage}{\linewidth}
    \captionof{table}{Continual steering over eight tasks on CALVIN (in imagination) and LIBERO (in the environment), with the Single-task and Current-task columns that Table~\ref{tab:sequence} omits. Success and Forgetting are the AVG and NBT of \citet{hu2026continualvlarl}.}
    \label{tab:app_sequence_full}
    % Continual steering on CALVIN and LIBERO, benchmarks side by side.
\setlength{\tabcolsep}{0.9pt}
\renewcommand{\arraystretch}{1.05}
{\fontsize{6}{7.2}\selectfont
\begin{tabular*}{\linewidth}{@{\extracolsep{\fill}}lccccc|ccccc@{}}
\toprule
& \multicolumn{5}{c|}{\textbf{CALVIN}}
& \multicolumn{5}{c}{\textbf{LIBERO}} \\
\cmidrule(lr){2-6}\cmidrule(lr){7-11}
\textbf{Method}
& Single-task$\uparrow$
& Current-task$\uparrow$
& Success$\uparrow$
& Retention$\uparrow$
& Forgetting$\downarrow$
& Single-task$\uparrow$
& Current-task$\uparrow$
& Success$\uparrow$
& Retention$\uparrow$
& Forgetting$\downarrow$ \\
\midrule
Base
& $63.7 \pm 6.5$
& $63.7 \pm 6.5$
& $63.7 \pm 6.5$
& $0.0$
& $0.0$
& $10.4 \pm 2.7$
& $10.4 \pm 2.7$
& $10.4 \pm 2.7$
& $0.0$
& $0.0$ \\

LoRA
& $67.1 \pm 4.9$
& $58.2 \pm 5.9$
& $11.7 \pm 1.9$
& $-52.0$
& $53.2$
& $59.2 \pm 4.5$
& $54.7 \pm 1.6$
& $1.0 \pm 1.6$
& $-41.6$
& $61.4$ \\

DSRL
& $\mathbf{81.8 \pm 2.9}$
& $65.0 \pm 5.6$
& $63.6 \pm 4.8$
& $-0.1$
& $\mathbf{1.6}$
& $\mathbf{91.3 \pm 1.0}$
& $\mathbf{85.4 \pm 0.0}$
& $16.9 \pm 5.2$
& $-30.8$
& $78.3$ \\

\hspace{0.5em}+ instruction KL
& $79.8 \pm 4.6$
& $74.0 \pm 7.5$
& $70.9 \pm 8.1$
& $+7.2$
& $3.6$
& $81.7 \pm 1.0$
& $78.6 \pm 10.9$
& $24.9 \pm 2.6$
& $-23.3$
& $61.4$ \\

\hspace{0.5em}+ first-obs KL
& $76.0 \pm 7.3$
& $76.5 \pm 1.6$
& $65.1 \pm 3.6$
& $+1.4$
& $12.9$
& $82.2 \pm 2.6$
& $82.3 \pm 6.2$
& $36.4 \pm 0.6$
& $-20.9$
& $52.5$ \\

\hspace{0.5em}+ replay KL
& $79.3 \pm 4.1$
& $79.1 \pm 3.5$
& $74.0 \pm 3.5$
& $+10.3$
& $5.8$
& $81.3 \pm 6.3$
& $84.0 \pm 6.6$
& $73.9 \pm 6.3$
& $\mathbf{+2.0}$
& $11.6$ \\

\textbf{D2R (Ours)}
& $79.8 \pm 3.8$
& $\mathbf{79.8 \pm 3.8}$
& $\mathbf{74.8 \pm 3.7}$
& $\mathbf{+11.1}$
& $5.7$
& $82.2 \pm 9.3$
& $82.1 \pm 2.7$
& $\mathbf{74.9 \pm 2.6}$
& $+1.6$
& $\mathbf{8.3}$ \\

\midrule

\textit{Per-task modules}
& $81.8 \pm 2.9$
& $81.8 \pm 2.9$
& $81.8 \pm 2.9$
& $+18.1$
& $0.0$
& $91.3 \pm 1.0$
& $91.3 \pm 1.0$
& $91.3 \pm 1.0$
& $+80.9$
& $0.0$ \\

\textit{Multi-task module}
& $67.3 \pm 3.9$
& $67.3 \pm 3.9$
& $67.3 \pm 3.9$
& $\times$
& $\times$
& $54.6 \pm 8.5$
& $54.6 \pm 8.5$
& $54.6 \pm 8.5$
& $\times$
& $\times$ \\

\bottomrule
\end{tabular*}}
\end{minipage}\end{center}

\subsection{Full Per-Task Results}
\label{sec:app_pertask}
\begin{figure}[h]
    \centering
    \input{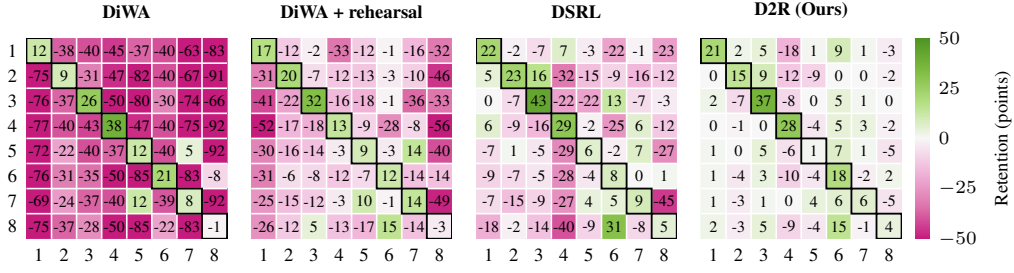}
    \caption{Retention in single-task adaptation on CALVIN for all steered and fine-tuned methods of Table~\ref{tab:calvin_results_v2}. Each cell shows the change in success relative to the base policy in points after adapting to the row task and evaluating on the column task, and outlined cells mark the adapted task. DiWA + rehearsal recovers most of DiWA's loss on the other tasks but remains below the base policy, DSRL drifts on several tasks, and D2R stays within a few points everywhere.}
    \label{fig:app_single_heatmaps}
\end{figure}

\begin{figure}[h]
    \centering
    \input{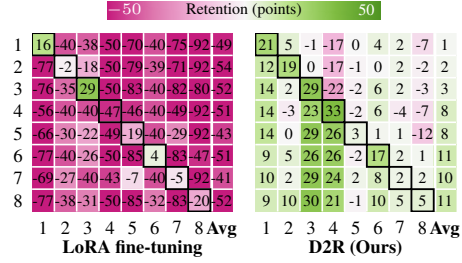}
    \caption{Continual steering on CALVIN, the sequence counterpart of Figure~\ref{fig:calvin_retention_heatmaps}. Change in success relative to the base policy after each stage (rows) on each task (columns) for LoRA fine-tuning and D2R, outlined cells mark the task learned at that stage, Avg is the mean over the eight tasks.}
    \label{fig:app_calvin_sequence}
\end{figure}

Every cell of Figures~\ref{fig:app_single_heatmaps} and~\ref{fig:app_calvin_sequence} is the retention
$S_{k,j}-S_{0,j}$ of \S\ref{sec:app_metrics}, the change in success on the evaluated task relative to the
frozen base policy. Green marks a gain, pink a loss, and white a system that ends where it started. This is
not the encoding of Figures~\ref{fig:calvin_retention_heatmaps} and~\ref{fig:franka} in the main text, whose
cells carry two different quantities on two scales, the success reached on the diagonal and the points lost
below it, and whose upper triangle is therefore empty. Here one signed quantity fills the whole matrix, which
separates two things those figures combine. A method that forgets heavily can still end above the base
policy, and a method that forgets nothing can end below it.

The two figures differ in what a row means. In Figure~\ref{fig:app_single_heatmaps} each row is an
independent adaptation to one task, so the rows are eight separate runs and carry no order. In
Figure~\ref{fig:app_calvin_sequence} each row is a stage of the sequence, so reading down a column shows how
one task fares as later tasks are learned, and the Avg column is the mean over the eight tasks.

\begin{figure}[h]
    \centering
    % Per-suite version (appendix). Data and styles come from results/libero_suite_retention_data.tex.
\input{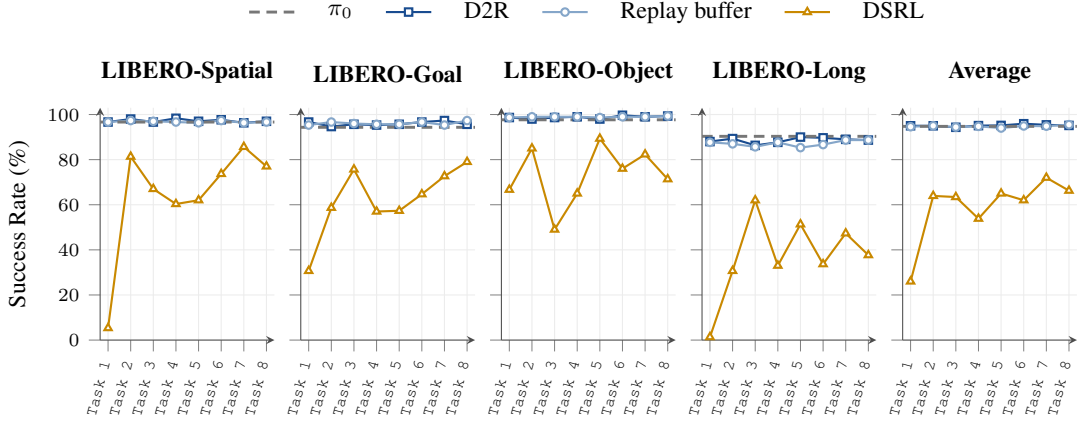}

\begin{tikzpicture}
\begin{groupplot}[
    group style={
        group size=5 by 1,
        horizontal sep=0.025\textwidth,
        y descriptions at=edge left
    },
    width=0.165\textwidth,
    height=0.22\textwidth,
    scale only axis,
    xmin=-0.35,
    xmax=7.35,
    ymin=0,
    ymax=103,
    xtick={0,1,2,3,4,5,6,7},
    xticklabels={
        \texttt{Task 1},
        \texttt{Task 2},
        \texttt{Task 3},
        \texttt{Task 4},
        \texttt{Task 5},
        \texttt{Task 6},
        \texttt{Task 7},
        \texttt{Task 8}
    },
    xticklabel style={
        rotate=75,
        anchor=east,
        font=\tiny
    },
    ytick={0,20,40,60,80,100},
    yticklabel style={font=\scriptsize},
    ylabel={Success Rate (\%)},
    ylabel style={font=\small},
    title style={font=\small\bfseries},
    axis lines=left,
    axis line style={draw=black!70},
    tick align=outside,
    grid=major,
    major grid style={draw=gray!15},
    legend cell align=left,
    legend style={
        draw=none,
        fill=none,
        font=\small,
        column sep=1em
    }
]

% ========================================================
% LIBERO-Spatial
% ========================================================
\nextgroupplot[
    title={LIBERO-Spatial},
    legend to name=liberoRetentionFiveLegend,
    legend columns=4
]

% Explicit legend order
\addlegendimage{base curve}
\addlegendentry{$\pi_0$}

\addlegendimage{imagined curve}
\addlegendentry{D2R}

\addlegendimage{real curve}
\addlegendentry{Replay buffer}

\addlegendimage{nokl curve}
\addlegendentry{DSRL}

\addplot[imagined curve,forget plot]
    table[x=stage,y=imagined] {\spatialdata};

\addplot[real curve,forget plot]
    table[x=stage,y=real] {\spatialdata};

\addplot[nokl curve,forget plot]
    table[x=stage,y=nokl] {\spatialdata};

% Draw the baseline last for visibility
\addplot[base curve,forget plot]
    coordinates {
        (-0.35,\piSpatial)
        (7.35,\piSpatial)
    };

% ========================================================
% LIBERO-Goal
% ========================================================
\nextgroupplot[title={LIBERO-Goal}]

\addplot[imagined curve]
    table[x=stage,y=imagined] {\goaldata};

\addplot[real curve]
    table[x=stage,y=real] {\goaldata};

\addplot[nokl curve]
    table[x=stage,y=nokl] {\goaldata};

\addplot[base curve]
    coordinates {
        (-0.35,\piGoal)
        (7.35,\piGoal)
    };

% ========================================================
% LIBERO-Object
% ========================================================
\nextgroupplot[title={LIBERO-Object}]

\addplot[imagined curve]
    table[x=stage,y=imagined] {\objectdata};

\addplot[real curve]
    table[x=stage,y=real] {\objectdata};

\addplot[nokl curve]
    table[x=stage,y=nokl] {\objectdata};

\addplot[base curve]
    coordinates {
        (-0.35,\piObject)
        (7.35,\piObject)
    };

% ========================================================
% LIBERO-Long
% ========================================================
\nextgroupplot[title={LIBERO-Long}]

\addplot[imagined curve]
    table[x=stage,y=imagined] {\longdata};

\addplot[real curve]
    table[x=stage,y=real] {\longdata};

\addplot[nokl curve]
    table[x=stage,y=nokl] {\longdata};

\addplot[base curve]
    coordinates {
        (-0.35,\piLong)
        (7.35,\piLong)
    };

% ========================================================
% Average across the four suites
% ========================================================
\nextgroupplot[title={Average}]

\addplot[imagined curve]
    table[x=stage,y=imagined] {\averagedata};

\addplot[real curve]
    table[x=stage,y=real] {\averagedata};

\addplot[nokl curve]
    table[x=stage,y=nokl] {\averagedata};

\addplot[base curve]
    coordinates {
        (-0.35,\piAverage)
        (7.35,\piAverage)
    };

\end{groupplot}

% Shared legend centered above all five panels
\path (group c1r1.north west) --
    node[
        midway,
        above=28pt,
        inner sep=0pt
    ]{\pgfplotslegendfromname{liberoRetentionFiveLegend}}
    (group c5r1.north east);

\end{tikzpicture}
    \caption{\textbf{Retention of $\pi_0$'s pretrained skills on LIBERO.} Success on each of the four suites
    $\pi_0$ was fine-tuned on after every stage of the eight-task sequence, with their mean in the last
    panel. The dashed line is the initial $\pi_0$, and the mean after the last stage is the LIBERO retention
    of Table~\ref{tab:sequence}.}
    \label{fig:libero-suite-retention}
\end{figure}

Figure~\ref{fig:libero-suite-retention} does the same for LIBERO, where the retained skills are
the four suites $\pi_0$ was fine-tuned on rather than individual tasks. Table~\ref{tab:sequence} reports only
their mean after the last stage, so the figure adds the trajectory, showing whether the pretrained skills
decay gradually or drop at one stage and whether some suites suffer more than others.

\subsection{Initialization of the Steering Module}
\label{sec:app_init}

At each stage the steering module $\piw_k$ can either continue from the module of the previous stage,
$\piw_{k-1}$, or start from scratch. Carrying the weights forward is the obvious choice, but fine-tuning the
same network through a sequence of stages costs plasticity \citep{dohare2024loss}, leaving the module worse
at learning each new task. Starting from scratch does not discard the previous stage, since the rehearsal
penalty of Eq.~\ref{eq:kl} still anchors $\piw_k$ to $\piw_{k-1}$ on the retained tasks, so behavior is
carried by the loss rather than by the weights.

Table~\ref{tab:incremental_ablations} compares the two with everything else held fixed. Reinitializing is
better on final success and on retention on both benchmarks, by $7.9$ points on each axis on CALVIN and by
$8.1$ and $0.4$ points on LIBERO, and it cuts forgetting on LIBERO from $18.6$ to $8.3$. Forgetting on CALVIN
is unchanged. We therefore reinitialize at every stage, as recorded in Table~\ref{tab:app_hparams}.

\begin{center}\begin{minipage}{\linewidth}
    \captionof{table}{Initialization of the steering module and sensitivity to task order. Above the rule, initializing $\piw_k$ from the previous stage $\piw_{k-1}$ against initializing it from scratch at every stage, both on the default order A. Below the rule, three further orders, all initialized from scratch.}
    \label{tab:incremental_ablations}

\setlength{\tabcolsep}{1.2pt}
\renewcommand{\arraystretch}{0.95}
{\fontsize{6}{7.2}\selectfont
\begin{tabular*}{\linewidth}{@{\extracolsep{\fill}}l|ccc|ccc@{}}
\toprule
& \multicolumn{3}{c|}{\textbf{CALVIN}}
& \multicolumn{3}{c}{\textbf{LIBERO}} \\
\cmidrule(lr){2-4}\cmidrule(lr){5-7}
\textbf{Setting}
& Success$\uparrow$ & Retention$\uparrow$ & Forgetting$\downarrow$
& Success$\uparrow$ & Retention$\uparrow$ & Forgetting$\downarrow$ \\
\midrule
Base & $63.7 \pm 6.5$ & $0.0$ & $0.0$ & $10.4 \pm 2.7$ & $0.0$ & $0.0$ \\
From previous & $66.9 \pm 3.0$ & $+3.2$ & 5.4 & $66.8 \pm 5.9$ & $+1.2$ & 18.6 \\
From scratch & $74.8 \pm 3.7$ & $+11.1$ & 5.7 & $\mathbf{74.9 \pm 2.6}$ & $\mathbf{+1.6}$ & $\mathbf{8.3}$ \\
\midrule
Order B & $74.6 \pm 2.0$ & $+10.9$ & 2.1 & -- & -- & -- \\
Order C & $75.9 \pm 7.1$ & $+12.2$ & 1.7 & -- & -- & -- \\
Order D & $\mathbf{78.9 \pm 1.9}$ & $\mathbf{+15.2}$ & $\mathbf{1.0}$ & -- & -- & -- \\
\bottomrule
\end{tabular*}}

\end{minipage}\end{center}

\subsection{Task Order Sensitivity}
\label{sec:app_task_order}

We evaluate four task orders on CALVIN. Order A is the default used throughout the paper, and orders
B, C and D are independent random permutations of the same eight tasks, with every other setting held fixed.

Table~\ref{tab:incremental_ablations} places all four within a few points of one another on every metric.
Final success spans $74.6\%$ to $78.9\%$, retention spans $+10.9$ to $+15.2$ points, and forgetting spans
$1.0$ to $5.7$ points. Within that spread the default order is not a favorable one, ranking third of the four
on success and on retention and last on forgetting, so the sequence reported in the main paper does not
flatter the method.
\subsection{Retention Coefficient $\lambda$}
\label{sec:app_lambda}

\begin{figure}[h]
    \centering
    \includegraphics[width=0.68\textwidth, trim=0 0 306.5bp 0, clip]{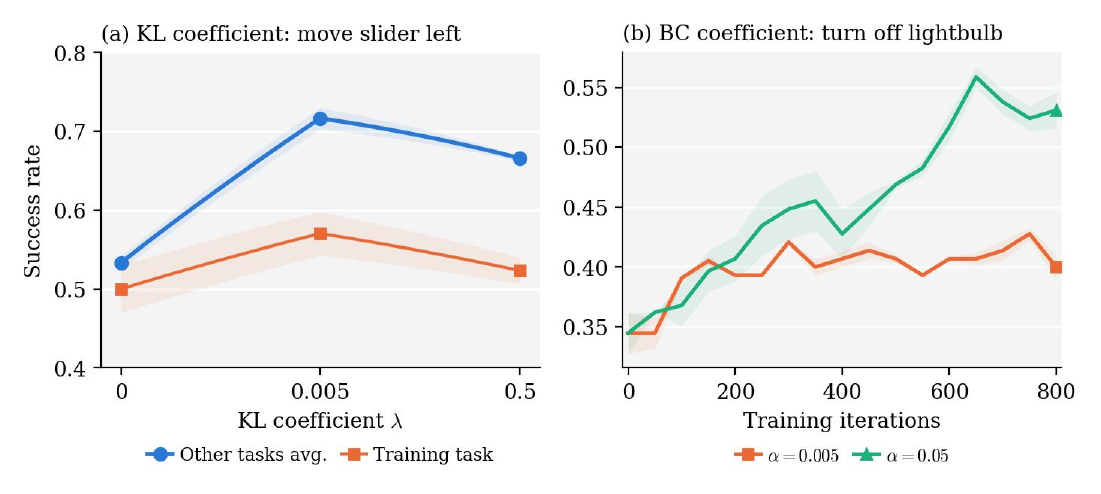}
    \caption{\textbf{Loss coefficients on CALVIN.}
    \textbf{(a)} Effect of the retention coefficient $\lambda$ on the success of the target task
    (\texttt{move\_slider\_left}) and the average success across previously learned tasks.
    \textbf{(b)} Effect of the behavior-cloning coefficient $\alpha$ on learning the target task
    (\texttt{turn\_off\_lightbulb}) over training.}
    \label{fig:app_coeff_sweeps}
\end{figure}

The retention coefficient $\lambda$ controls the trade-off between learning the current task and preserving performance on previously learned tasks. As shown in Figure~\ref{fig:app_coeff_sweeps}(a) on the \texttt{move\_slider\_left} task from CALVIN benchmark, setting $\lambda=0$ leads to lower retention, whereas an excessively large value limits adaptation to the current task. We use $\lambda=0.005$, which provides the best balance between current task success and average retained success.

\subsection{Behavior Cloning Coefficient $\alpha$}
\label{sec:app_alpha}

The behavior cloning coefficient $\alpha$ controls the contribution of the behavior cloning
objective during steering policy optimization. Figure~\ref{fig:app_coeff_sweeps}(b) compares $\alpha=0.005$
and $\alpha=0.05$ on the \texttt{turn\_off\_lightbulb} task from CALVIN benchmark. A small coefficient can
result in shortcut learning, world model exploitation, or reward hacking, which leads to poor performance
when tested on the real environment. A balanced coefficient results in faster learning and a higher final
success rate, so we use $\alpha=0.05$ in our experiments. A larger value could instead reduce the
exploration of the policy and limit it to the base policy.
\subsection{World Model Faithfulness Check}
\label{sec:app_faithfulness_check}

\begin{figure}[h]
    \centering
    \includegraphics[width=0.55\textwidth]{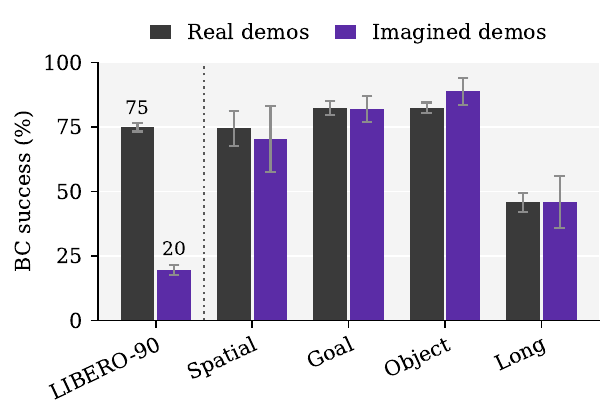}
    \caption{\textbf{Faithfulness of A2World on LIBERO.} Success of behavior cloning on real demonstrations
    against A2World-imagined replays of the same demonstrations. The two agree on the four suites A2World
    was fine-tuned on and diverge on the LIBERO-90 tasks of the sequence.}
    \label{fig:app_faithfulness_panel}
\end{figure}

Figure~\ref{fig:app_faithfulness_panel} shows that A2World cannot serve as a learning simulator for
the LIBERO-90 tasks, which lie outside its training data, while remaining faithful on the four suites it was
fine-tuned on. For each suite we train two behavior-cloning policies with the same architecture and
hyperparameters and different training data. The first is trained on 50 real expert demonstrations from that
suite. The second is trained on 50 imagined demonstrations, obtained by initializing A2World with the
initial scene of each expert demonstration and rolling it out under the corresponding expert action
sequence. If the world model were faithful, the two policies would reach comparable success.

On the four suites they do. Cloning from imagined replays reaches $70.3\%$ against $74.5\%$ on Spatial,
$82.0\%$ against $82.5\%$ on Goal, $88.9\%$ against $82.4\%$ on Object, and $46.0\%$ against $45.8\%$ on
Long. On the eight LIBERO-90 tasks of the sequence the gap is wide, $19.6\%$ against $74.9\%$. A2World
reproduces the dynamics it was trained on and not those of the target tasks, so a steering module optimized
inside it would exploit its prediction errors rather than learn the task, which is why LIBERO adapts in the
environment and uses the world model only for rehearsal.

The BC policy is a Gaussian policy consisting of four convolutional layers, text conditioning with FiLM fusion, and three multilayer perceptron (MLP) layers that predict the mean and standard deviation of the action distribution. The policy predicts action chunks of length 20 and is trained for 100 epochs using a negative log-likelihood objective, a batch size of 256, Adam optimizer, and a learning rate of $10^{-4}$. We evaluate both policies on each LIBERO suite using 30 episodes per task and three different evaluation seeds.

\subsection{Qualitative Imagined Rehearsal Rollouts}

\begin{figure*}[!ht]
    \centering
    \includegraphics[width=\textwidth]{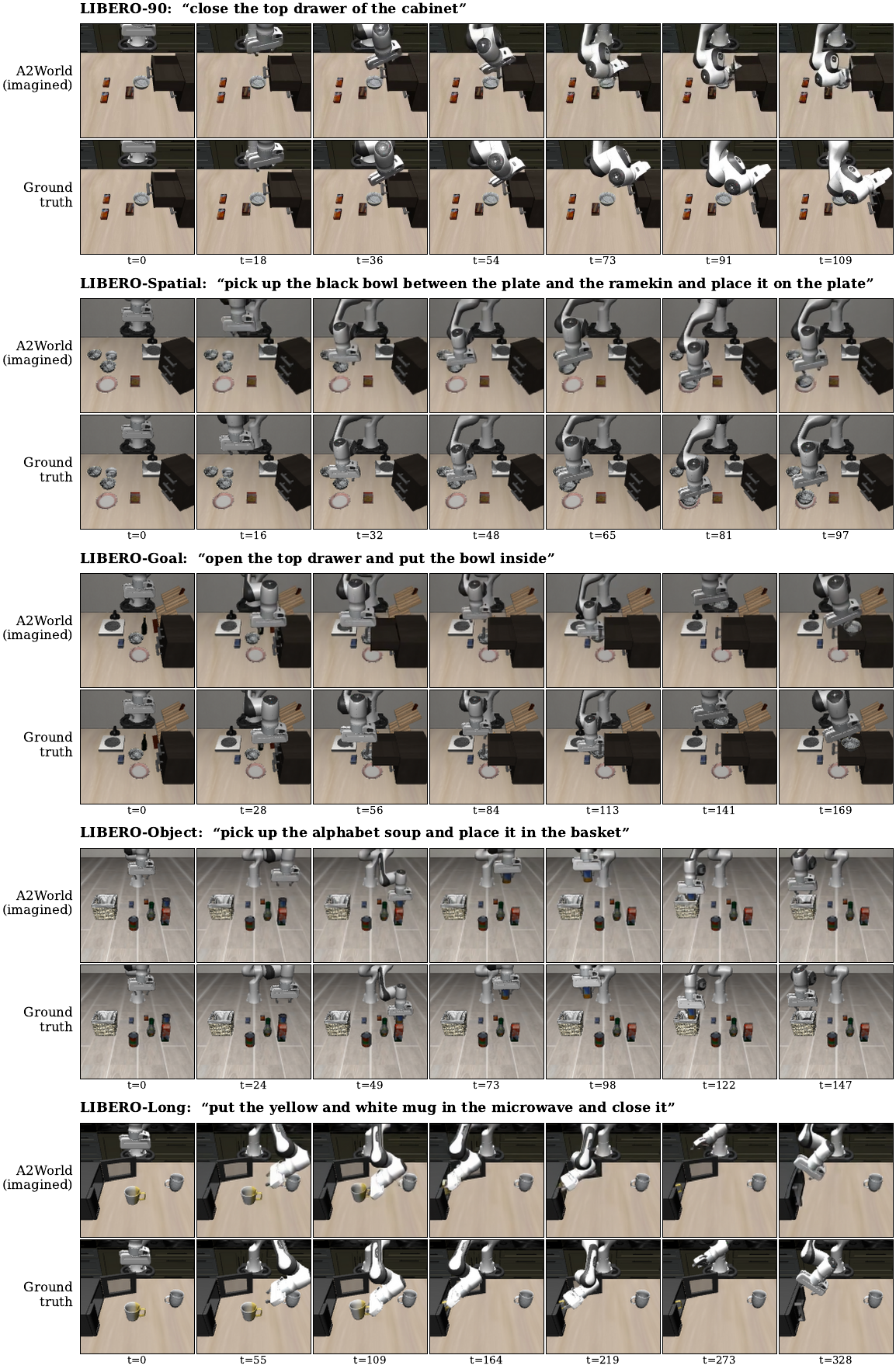}
    \caption{
        Qualitative comparison between trajectories imagined by A2World and the corresponding ground-truth trajectories across the LIBERO suites. Each pair of rows shows temporally aligned frames from an A2World rollout and the associated ground-truth demonstration.
    }
    \label{fig:a2world_vs_gt_libero}
\end{figure*}

We provide qualitative comparisons between imagined rehearsal rollouts and their corresponding ground-truth trajectories for the world models used in our experiments. Figure~\ref{fig:a2world_vs_gt_libero} shows A2World rollouts across the LIBERO suites. Although A2World does not consistently generate complete and successful task trajectories on LIBERO-90, it captures the environment dynamics and produces plausible transitions between different states. These rollouts therefore remain useful for rehearsal, where the objective is to preserve the steering policy's behavior on previously learned tasks rather than to learn a new task entirely from imagination.

\begin{figure*}[!ht]
    \centering
    \includegraphics[width=\textwidth]{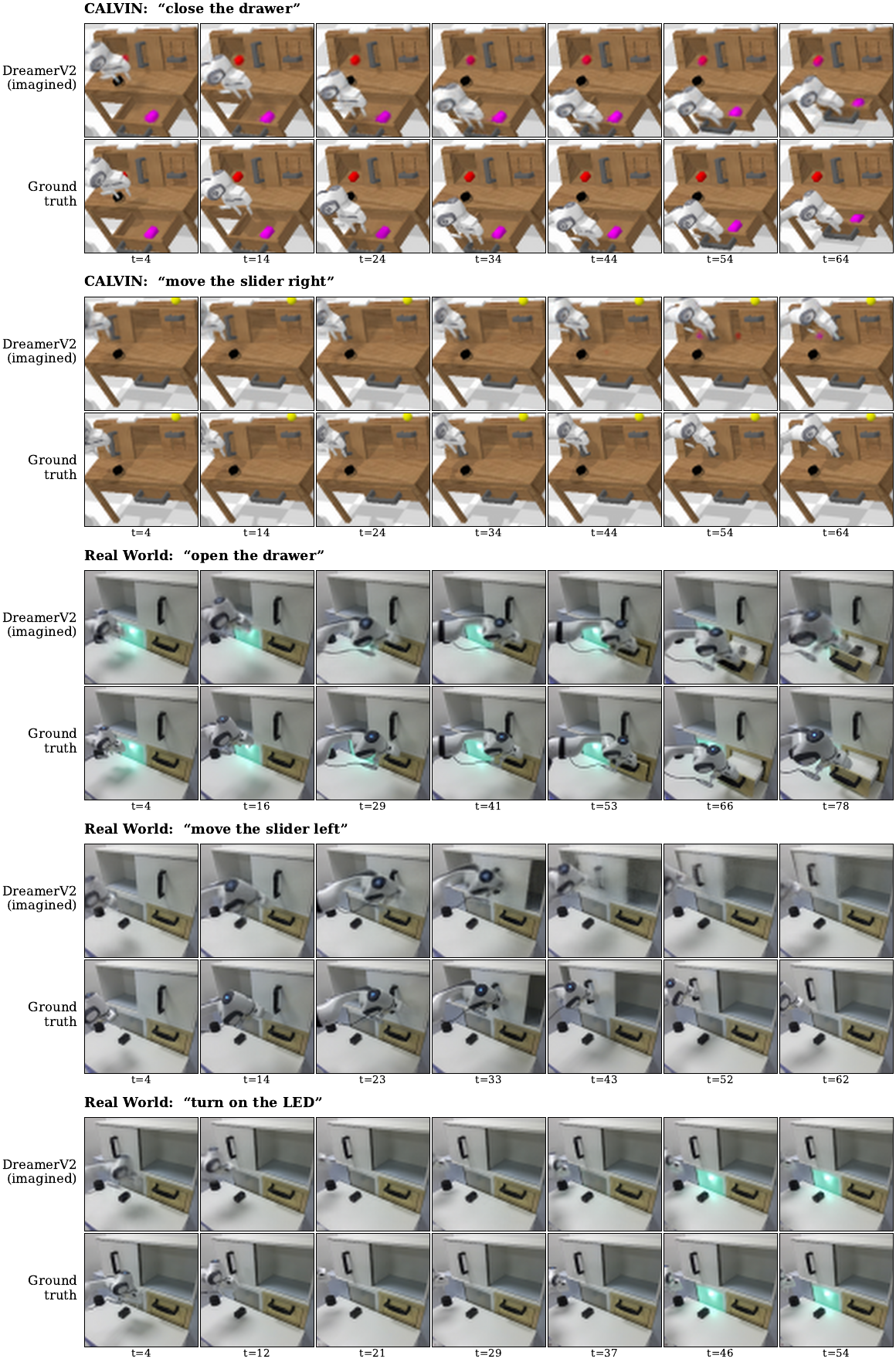}
    \caption{
        Qualitative comparison between trajectories imagined by DreamerV2 and the corresponding ground-truth trajectories across the CALVIN benchmark and Real-world experiment. Each pair of rows shows temporally aligned frames from an imagined rollout and the associated ground-truth demonstration.
    }
    \label{fig:dreamer_vs_gt}
\end{figure*}

Figure~\ref{fig:dreamer_vs_gt} presents analogous results for the DreamerV2 world model used in the CALVIN and real-world experiments. Compared with A2World on the unseen LIBERO-90 tasks, DreamerV2 produces imagined trajectories that more closely follow the corresponding ground-truth task progression. This higher fidelity enables DreamerV2 to serve both as a simulator for learning new tasks in imagination and as a generative memory for rehearsing previously learned tasks.

\end{document}